\documentclass[10pt]{article} %
\usepackage[accepted]{tmlr}

\usepackage{hyperref}
\usepackage{url}

\title{\vspace{-0.7cm}Differentially Private Synthetic Data via APIs 3:\\ Using Simulators Instead of Foundation Models}

\author{\name Zinan Lin \email zinanlin@microsoft.com \\
      \addr Microsoft Research\\
      Redmond, WA, USA
      \AND
        \name Tadas Baltrusaitis \email tadas.baltrusaitis@microsoft.com \\
      \addr Microsoft\\
      Cambridge, UK
      \AND
      \name Wenyu Wang  \email wenyuu.wang@gmail.com \\
      \addr Redmond, WA, USA
      \AND
      \name Sergey Yekhanin \email yekhanin@microsoft.com\\
      \addr Microsoft Research\\
      Redmond, WA, USA}

\def\month{08}  %
\def\year{2026} %
\def\openreview{\url{https://openreview.net/forum?id=GMoM2QpYuz}} %

\usepackage{etoc}

\usepackage{titletoc}
\usepackage{scalefnt}
\usepackage{gensymb}
\usepackage{placeins}

\usepackage{booktabs}

\usepackage{lipsum}  
\usepackage{wrapfig}
\usepackage{nicefrac}
\usepackage{hyperref}
\hypersetup{
    colorlinks=true,
    linkcolor=blue,
    filecolor=magenta,      
    urlcolor=cyan
}
\usepackage{mathtools}
\usepackage{amsmath,amsfonts}
\usepackage{amssymb}
\usepackage[ruled,vlined]{algorithm2e}
\usepackage{algorithmic}
\usepackage{makecell}
\usepackage{diagbox}
\usepackage{multirow}
\usepackage{textgreek}

\usepackage{listings}

\usepackage{caption}
\usepackage{subcaption}

\usepackage{amsthm}

\usepackage[nameinlink,capitalize]{cleveref}

\crefname{section}{\S}{\S}
\Crefname{section}{\S}{\S}
\crefname{appendix}{App.}{Apps.}
\Crefname{appendix}{App.}{Apps.}
\crefname{theorem}{Thm.}{Thms.}
\Crefname{theorem}{Thm.}{Thms.}
\crefname{proposition}{Prop.}{Props.}
\Crefname{proposition}{Prop.}{Props.}

\crefname{table}{Tab.}{Tabs.}
\Crefname{table}{Tab.}{Tabs.}

\crefname{algorithm}{Alg.}{Algs.}
\Crefname{algorithm}{Alg.}{Algs.}
\crefname{assumption}{Asm.}{Asms.}
\Crefname{assumption}{Asm.}{Asms.}
\crefname{mechanism}{Mech.}{Mechs.}
\Crefname{mechanism}{Mech.}{Mechs.}

\crefformat{footnote}{#2\footnotemark[#1]#3}
\crefmultiformat{footnote}{#2\footnotemark[#1]#3}%
{\textsuperscript{,}#2\footnotemark[#1]#3}{\textsuperscript{,}#2\footnotemark[#1]#3}{\textsuperscript{,}#2\footnotemark[#1]#3}

\makeatletter
\newcommand\footnoteref[1]{\protected@xdef\@thefnmark{\ref{#1}}\@footnotemark}
\makeatother

\newcommand{\myparatightestn}[1]{ \noindent\textbf{{#1}}}

\newcounter{packednmbr}

\newenvironment{packeditemize}{\begin{list}{$\bullet$}{\setlength{\itemsep}{0.5pt}\addtolength{\labelwidth}{-4pt}\setlength{\leftmargin}{\labelwidth}\addtolength{\leftmargin}{-2pt}\setlength{\listparindent}{\parindent}\setlength{\parsep}{1pt}\setlength{\topsep}{0pt}}}{\end{list}}

\NewDocumentCommand{\codeword}{v}{%
\texttt{\textcolor{blue}{#1}}%
}

\newcommand{\revisiontmlr}[1]{{#1}}

\newcommand{\size}[2]{{\fontsize{#1}{0}\selectfont#2}}

 \newcommand\blfootnote[1]{%
  \begingroup
  \renewcommand\thefootnote{}\footnote{#1}%
  \addtocounter{footnote}{-1}%
  \endgroup
}

\newcommand{\classavail}{\textbf{ClassAvail}}

\newcommand{\classunavail}{\textbf{ClassUnavail}}

\newcommand{\simpe}{{\scalefont{0.9}\textsf{\textcolor{orange}{Sim-PE}}}}

\newcommand{\kubric}{\textsc{Kubric}}
\newcommand{\teapot}{\textsc{3D Teapot}}
\newcommand{\pythonavatar}{\textsc{python-avatar}}
\newcommand{\metasim}{\textsc{Meta-Sim}}
\newcommand{\facesynthetics}{\textsc{Face Synthetics}}
\newcommand{\digiface}{\textsc{DigiFace-1M}}

\newcommand{\privateevolution}{{\scalefont{0.9}\textsf{\textcolor{black}{Private Evolution}}}}
\newcommand{\pe}{{\scalefont{0.9}\textcolor{black}{\textsf{PE}}}}
\newcommand{\dpvotingname}{DP Nearest Neighbors Histogram}

\newcommand{\celeba}{CelebA}
\newcommand{\mnist}{MNIST}
\newcommand{\cifar}{CIFAR10}
\newcommand{\imagenet}{ImageNet}

\newcommand{\openreviewdataset}{OpenReview}
\newcommand{\cartpole}{CartPole}

\newcommand{\numiterations}{T}
\newcommand{\numgensamples}{N_\syn}
\newcommand{\numprisamples}{N_\priv}
\newcommand{\privatesampleset}{S_\priv}
\newcommand{\privatesampleclassset}{C}
\newcommand{\generatedsampleset}{S_\syn}
\newcommand{\noisemultiplier}{\sigma}
\newcommand{\threshold}{H}
\newcommand{\alg}{\calM}

\newcommand{\randomsampleapiname}{\size{9}{\textsf{RANDOM\_API}}}
\newcommand{\randomsampleapi}[1]{\randomsampleapiname{}\bra{#1}}

\newcommand{\samplevariationapiname}{\size{9}{\textsf{VARIATION\_API}}}
\newcommand{\samplevariationapi}[1]{\samplevariationapiname{}\bra{#1}}

\newcommand{\dpvotingfunctionname}{\size{9}{\textsf{DP\_NN\_HISTOGRAM}}}
\newcommand{\dpvotingfunction}[1]{\dpvotingfunctionname{}\bra{#1}}

\newcommand{\distancefunctionname}{d}
\newcommand{\distancefunction}[1]{\distancefunctionname{}\bra{#1}}

\newcommand{\normaldistribution}[2]{\calN\bra{#1,#2}}

\newcommand{\priv}{\mathrm{priv}}
\newcommand{\syn}{\mathrm{syn}}

\newcommand{\probnotation}{\mathbb{P}}
\newcommand{\probof}[1]{\probnotation\bra{#1}}

\newcommand{\bra}[1]{\left( #1 \right)}
\newcommand{\brb}[1]{\left[ #1 \right]}

\newcommand{\brc}[1]{\left\{ #1 \right\}}
\newcommand{\brd}[1]{\left| #1 \right|}

\newcommand{\calD}{\mathcal{D}}

\newcommand{\calM}{\mathcal{M}}
\newcommand{\calN}{\mathcal{N}}

\newcommand{\calP}{\mathcal{P}}

\newcommand{\cate}{\xi}
\newcommand{\nume}{\phi}
\newcommand{\numcate}{p}
\newcommand{\numnume}{q}
\newcommand{\cateset}{\Xi}
\newcommand{\numeset}{\Phi}
\newcommand{\simulator}{\mathcal{S}}
\newcommand{\simulatorfun}[1]{\simulator\bra{#1}}

\newcommand{\uniform}[1]{\text{Uniform}\bra{#1}}

\newcommand{\catevariationdegree}{\beta}
\newcommand{\numevariationdegree}{\alpha}
\newcommand{\nnvariationdegree}{\gamma}

\newcommand{\simdata}{z}
\newcommand{\numsimdata}{m}
\newcommand{\simdataset}{S_{\text{sim}}}

\newcommand{\numclusters}{N_\text{cluster}}
\newcommand{\clustercenter}{w}

\newcommand{\nnsample}{q}%
\renewcommand{\cite}{\citep}

\begin{document}

\maketitle

\vspace{-0.4cm}
\begin{abstract}
Differentially private (DP) synthetic data enables the use of sensitive datasets while providing strong privacy guarantees. Recently, \privateevolution{} (\pe{}) has emerged as a promising framework for generating DP synthetic data using only inference APIs of foundation models. However, suitable foundation models are not always available for every private data domain.
In this paper, we show that \pe{} is more general than previously understood and can incorporate APIs beyond foundation models. In particular, we show that state-of-the-art \emph{non–neural-network data synthesizers}, including computer graphics-based image generators and physics simulation tools (which we refer to as \emph{simulators}), integrate naturally into \pe{}. We call the resulting method \simpe{}.
\textbf{\simpe{} broadens the applicability of \pe{} while improving performance}. For image synthesis, \simpe{} improves downstream classification accuracy by up to $3\times$, reduces FID by up to 80\%, and is significantly more efficient. We further show that simulators and foundation models can be seamlessly combined within \simpe{} for additional gains.
Beyond performance improvements, \textbf{\simpe{} enables new DP applications that are difficult or impossible for existing methods}, including generating DP control signals whose execution reproduces the distribution of private observations.
The code is open-sourced in the \privateevolution{} Python library: \url{https://github.com/microsoft/DPSDA}.
\end{abstract}

\blfootnote{$^\dagger$ Main updates in arXiv v4 (TMLR camera-ready): New experiments extending the method to additional application domains, including DP generation of control signals and networking data; improvements to the writing and presentation.}

\vspace{-1.cm}
\section{Introduction}
\label{sec:intro}
Leaking sensitive user information is a major concern in data-driven applications. A common solution is to generate differentially private (DP) \cite{dwork2006calibrating} synthetic data that resembles the original while ensuring strong privacy guarantees. Such data can substitute the original in tasks like model fine-tuning, statistical analysis, and data sharing, while preserving user privacy \cite{bowen2019comparative,lin2022data,tao2021benchmarking,hu2024sok}.

\privateevolution{} (\pe{}) \cite{lin2023differentially,xie2024differentially,tran2026differentially} has recently emerged as a promising method for DP data synthesis, adopted in Microsoft and Apple \cite{apple_pe,microsoft_pe}. 
It begins by probing foundation models to produce random samples, then iteratively selects those most similar to private data and uses the model to generate more like them.
Unlike prior %
methods that fine-tune open-source models, \pe{} relies solely on model inference--making it up to 66$\times$ faster \cite{xie2024differentially}. 
More importantly, this allows \pe{} to easily leverage cutting-edge blackbox models like GPT-5 \cite{singh2025openai}, %
achieving SoTA performance on multiple image, text, and tabular benchmarks \cite{lin2023differentially,xie2024differentially,tran2026differentially,zou2025contrastive,hou2025private,wang2025synthesize,wang2025struct,chien2026maple}.\footnote{We refer readers to \url{https://zinanlin.me/private-evolution-papers/} for an up-to-date list of papers on \privateevolution{}.
}

However, \pe{} relies on foundation models suited to the private data domain, which may not always be available. When the model's distribution significantly differs from the private data, \pe{}'s performance lags far behind training-based methods \cite{dpimagebench}.

To address this question, we note that in the traditional synthetic data field—where private data is \emph{not} involved—\emph{non-neural-network data synthesizers} remain widely used, especially in domains where foundation models struggle. Examples include computer graphics-based renders for images, videos, and 3D data (e.g., Blender \cite{blender} and Unreal \cite{unreal}) and physics-based simulators for robotics data (e.g., Gym/Gymnasium \cite{towers2024gymnasium,gym}, Genesis \cite{Genesis}). %
For brevity, we refer to these tools as \emph{simulators}. While these simulators have been successful, their applications in \emph{DP} data synthesis remain underexplored. This is understandable, as adapting these simulators to fit private data in a DP fashion requires non-trivial, case-by-case modifications.
Our key insight is that \pe{} only requires two APIs: \randomsampleapiname{} that generates random samples and \samplevariationapiname{} that generates samples similar to the given one. These APIs do not have to come from foundation models. Thus, we ask: 
\emph{Can \pe{} use simulators in place of models?}
If viable, this approach could greatly expand \pe{}'s capabilities and unlock the potential of a wide range of %
simulators for DP data synthesis.

\begin{figure}[t]
    \centering
    \includegraphics[width=0.4\linewidth]{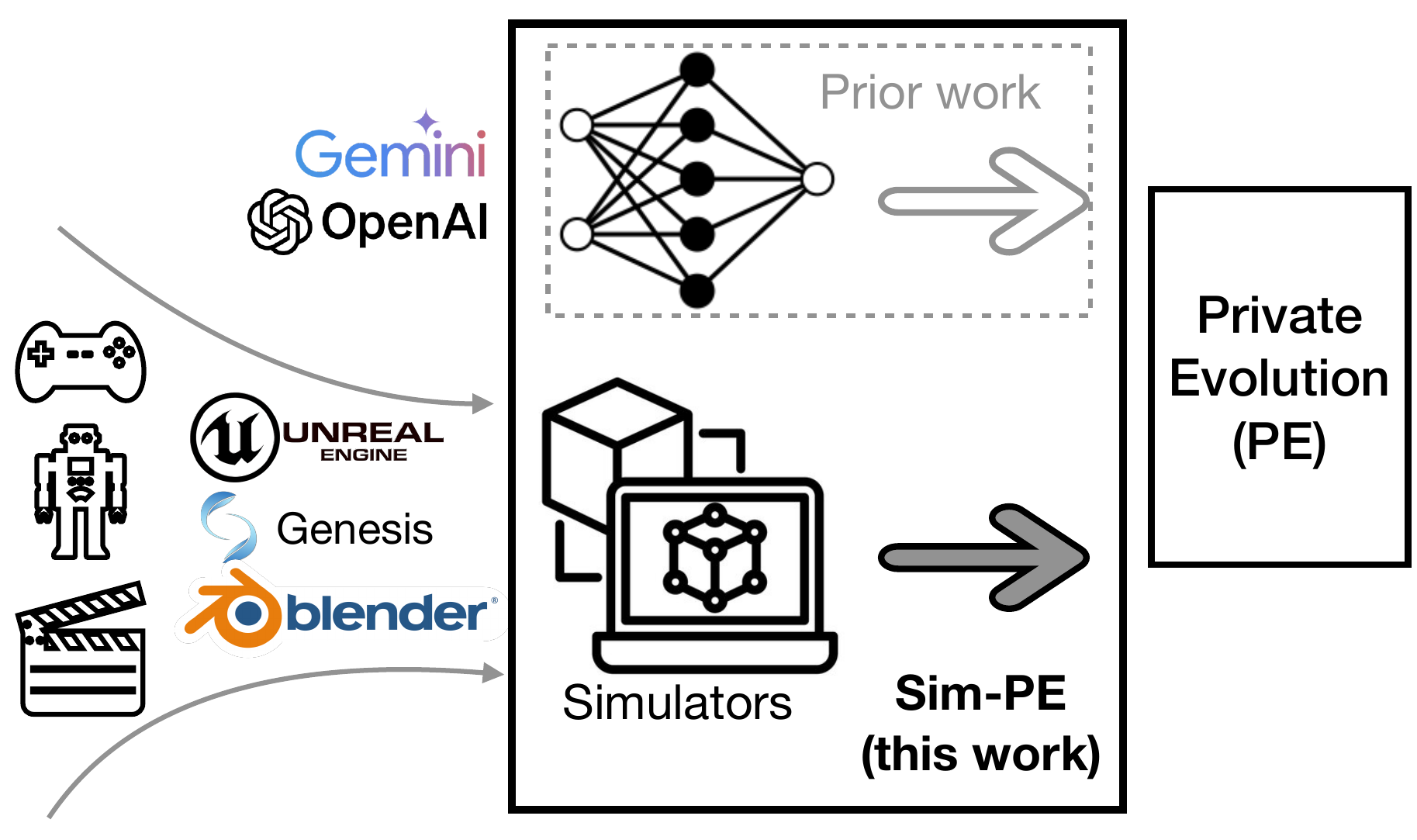}
    \caption{Unlike prior \pe{} work that relies solely on foundation models, we show that \pe{} is also compatible with non-neural-network data synthesis tools, which we call \emph{simulators}. This greatly broadens \pe{}'s applicability, enables SoTA simulators for DP data synthesis, and unlocks new DP applications. }
    \label{fig:overview}
\end{figure}

In this paper, we propose \simpe{} (\cref{fig:overview}) to exploit this potential. %
While the underlying idea is conceptually simple, its correct realization requires substantial effort. Through systematic exploration, we arrive at the following design, which distinguishes between two types of simulator access:
\textbf{(1) The simulator is accessible.} We define \randomsampleapiname{} as generating a sample with random simulator parameters, and \samplevariationapiname{} as slightly perturbing the simulator parameters of the given sample.
\textbf{(2) The simulator is inaccessible—only its generated data is released.} This scenario is quite common \cite{wood2021fake,bae2023digiface}, especially when simulator assets are proprietary \cite{kar2019meta,devaranjan2020meta}. In this case, we define \randomsampleapiname{} as randomly selecting a sample from the dataset, and \samplevariationapiname{} as randomly selecting a nearest neighbor of the given sample.
We demonstrate that with suitable simulators, \simpe{} can outperform \pe{} with models. Moreover, \simpe{} enables new applications that are difficult or infeasible for prior methods.
Our key contributions are:
\begin{packeditemize} 
    \item \textbf{Advancing \pe{}.} \textbf{(1)} We discover that \pe{} can leverage tools beyond foundation models and propose \simpe{}, which integrates simulators into \pe{}. %
    \revisiontmlr{This integration alleviates two key limitations of PE. \textbf{(a) Improved performance under large distribution shift.} \pe{} performs poorly when the private data distribution deviates significantly from the pretraining data of foundation models. By incorporating simulators, \simpe{} achieves substantially better performance in such settings.}
    For example, on the \mnist{} dataset with $\epsilon=1$, downstream classification accuracy improves to 89.1\%, compared to 27.9\% with the original \pe{}. \revisiontmlr{\textbf{(b) Improved efficiency.} As an additional benefit, \simpe{} reduces the high cost of repeated foundation model API calls in \pe{} by offloading computation to potentially cheaper simulators. For example, on \mnist{}, \simpe{} running on CPU is 80$\times$ faster than \pe{} on GPU, while also benefiting from the lower cost of CPU computation. }\\
    \textbf{(2)} We further introduce a unified framework that allows \emph{models and simulators to be used interchangeably}, enabling \pe{} to leverage the complementary strengths of both and yielding additional performance gains.
    \item \textbf{Unlocking new DP applications.} We show that \simpe{} supports applications in which the generation target and the private observations lie in \emph{different} spaces, such as \emph{generating DP control signals whose execution in a robotics simulator reproduces the distribution of private observations}. \textbf{This setting is fundamentally challenging for prior training-based methods} for two reasons: (1) they are designed for scenarios in which the generation target and observations reside in the same space, and (2) they are gradient-based, making them incompatible with simulators through which backpropagation is not possible.
\end{packeditemize}

\section{Preliminaries and Motivation}
\label{sec:background}
\subsection{Preliminaries}
\label{sec:preliminaries}

\myparatightestn{Synthetic data} refers to ``fake'' data generated by models or software for various applications, including data augmentation, model training, and software testing \cite{lin2022data}.
While neural-network-based generative models such as GANs \cite{goodfellow2020generative}, diffusion models \cite{sohl2015deep}, and auto-regressive models \cite{openai2023gpt4,liu2024distilled,liu2025distilled} are widely used, non-neural-network tools remain SoTA in many applications. For example, ns \cite{issariyakul2009introduction,riley2010ns} can simulate networks and generate network packets based on network configurations. Blender \cite{blender}, given 3D models and lighting configurations, can render images and videos, and is extensively used in movie production. \textbf{In this paper, we refer to these tools as \emph{simulators}.} See \cref{app:simulators} for a discussion on the continued importance of simulators, even in the era of large foundation models.

\myparatightestn{DP synthetic data} requires the synthetic data to be close to a given private dataset, while having a differential privacy (DP) \cite{dwork2006calibrating} guarantee. Formally, a mechanism $\alg$ is $(\epsilon,\delta)$-DP if for any two neighboring datasets $\calD$ and $\calD'$ (i.e., $\calD'$ has one extra entry compared to $\calD$ or vice versa) and for any set $S$ of outputs of $\alg$, we have 
$\probof{\alg\bra{\calD}\in S}\leq e^{\epsilon}\probof{\alg\bra{\calD'}\in S} + \delta$. 
Current SoTA DP image and text synthesis methods %
typically requires neural network 
training \cite{lin2020using,beaulieu2019privacy,dockhorn2022differentially,yin2022practical,yu2021differentially,he2022exploring,li2021large,ghalebikesabi2023differentially,yue2022synthetic,jordon2019pate,harder2022differentially,harder2021dp,vinaroz2022hermite,cao2021don,chen2022dpgen}.

\myparatightestn{\privateevolution{} (\pe{})} \cite{lin2023differentially,xie2024differentially,tran2026differentially} is a recent training-free framework for DP data synthesis.
Therefore, unlike prior training-based methods, \pe{} can leverage state-of-the-art models even when they are accessible only via APIs (e.g., GPT-5), while also being more computationally efficient \cite{lin2023differentially,xie2024differentially}. 
In more detail, 
\textbf{(1) Performance.} 
\pe{} and its follow-up methods have achieved state-of-the-art performance on several \emph{image}, \emph{text}, and \emph{tabular} benchmarks \cite{lin2023differentially,xie2024differentially,tran2026differentially,wang2025struct,chien2026maple}.  
\emph{When using comparable open-source pre-trained models}, \pe{} achieves an \emph{image} quality score of FID $\leq 7.9$ on \cifar{} with a privacy cost of $\epsilon=0.67$,  
representing a significant improvement over the previous state-of-the-art, which required $\epsilon=32$ \cite{lin2023differentially}.    
\emph{By leveraging state-of-the-art models behind APIs---where training-based methods are not applicable---}\pe{} further improves performance,  
outperforming all prior approaches in downstream \emph{text} classification accuracy on the \openreviewdataset{} dataset \cite{xie2024differentially}.  
\textbf{(2) Efficiency.}
Furthermore, since \pe{} only requires model inference, it can be up to 66$\times$ more efficient than training-based methods for DP \emph{text} generation \cite{xie2024differentially}.
\textbf{(3) Broader applications.}
Additionally, \pe{} can be applied to \emph{federated learning} to shift model training from devices to central servers in a differentially private and more efficient manner \cite{hou2024pre,zou2025contrastive,hou2025private}. 
While \pe{} itself does not require model training, it can also be extended to fine-tune model weights via reinforcement \cite{xu2026dp,hou2025private} or supervised \cite{zou2026pesgd} fine-tuning.
\pe{} has also been adapted for \emph{DP clustering} \cite{humphries2026pe}.
\pe{} has been adopted by some of the largest IT companies, including Microsoft and Apple \cite{apple_pe,microsoft_pe}.

\subsection{Motivation}

While \pe{} %
is promising, its performance significantly drops when there is a large distribution shift between the private data and the foundation model’s pre-trained data \cite{dpimagebench}. For example, using \mnist{} \cite{lecun1998mnist} as private data, training a downstream digit classifier (10 classes) on DP synthetic data generated by \pe{} with $\epsilon=1$—via a foundation model pre-trained on \imagenet{}—achieves only 27.9\% accuracy \cite{dpimagebench}. Because suitable foundation models may not exist for every domain, this limitation restricts \pe{}’s applicability in practice. Extending \pe{} to leverage simulators can substantially broaden its scope.

More broadly, simulators cannot be replaced by foundation models for data synthesis in many domains (\cref{sec:preliminaries,app:simulators}), yet many SoTA DP synthetic data methods rely heavily on (training) models and cannot leverage simulators. By extending \pe{} to simulators, we aim to unlock their potential for DP data synthesis and enable new applications infeasible for prior training-based methods.

\section{\simpe{}: Private Evolution with Simulators}

While the idea is conceptually simple, its implementation requires careful design. We first present an overview of the \simpe{} algorithm (\cref{sec:method_overview}), and then address three key questions: (1) How can \pe{} be adapted to simulators (\cref{sec:method_API})? (2) Can the strengths of models and simulators be leveraged jointly (\cref{sec:method_simulator_and_model})? (3) Beyond performance improvements, what unique capabilities does \simpe{} provide (\cref{sec:method_new_application})?

\subsection{Overview}
\label{sec:method_overview}

\myparatightestn{Key insight.} \pe{} requires a data generation backend that supports (1) \randomsampleapiname{}, which generates a random sample (e.g., a random bird image), and (2) \samplevariationapiname{}, which produces slight variations of a given sample (e.g., a similar bird image).  
Prior work on \pe{} \cite{lin2023differentially,xie2024differentially,hou2024pre,zou2025contrastive,hou2025private,swanberg2025api,wang2025synthesize,wang2025struct} has exclusively used foundation models to implement these APIs.  
Our key insight is that \emph{these APIs do not have to be powered by foundation models}: traditional data synthesizers that do not rely on neural networks—referred to as \emph{simulators}—can also be used to implement \randomsampleapiname{} and \samplevariationapiname{}.  

\myparatightestn{Algorithm overview.} Here, we describe the full workflow of \simpe{}. For completeness, the pseudocode is attached in \cref{app:pe}. 
We first use \randomsampleapiname{} to generate an initial set of random samples (\cref{line:initial}). Then, we iteratively refine these samples using the private data. In each iteration:
\begin{packeditemize}
    \item Each private sample casts a vote for its closest synthetic sample. This yields a histogram (denoted \dpvotingfunctionname{}) reflecting how well each synthetic sample aligns with the private data (\cref{line:voting}). To ensure differential privacy, Gaussian noise is added to it. %
    \item We sample synthetic data according to the noisy histogram, giving higher likelihood to those samples that align more closely with private data (\cref{line:normalization,line:draw}).
    \item We apply \samplevariationapiname{} to the drawn samples to generate additional variants (\cref{line:variation}). These samples become the initialization of the next iteration.
\end{packeditemize}
Final-iteration samples form the DP synthetic dataset.

\myparatightestn{Theoretical analysis.} \simpe{} provides strict DP guarantees with respect to the private dataset (including labels, if available) and guarantees convergence under assumptions in \citet{lin2023differentially}. We defer the privacy and convergence analysis, adapted from \pe{} \citep{lin2023differentially}, to \cref{app:theoretical}. %

\subsection{Simulator Adaptation in \simpe{}}
\label{sec:method_API}

Given the setup in \cref{sec:method_overview}, the core challenge is to design \randomsampleapiname{} and \samplevariationapiname{} for simulators. In practice, simulators expose different levels of access.

\myparatightestn{Simulator code access.} Some simulators are open-sourced. Examples include \kubric{} \cite{greff2021kubric}, a Blender-based renderer for multi-object images/videos; \teapot{} \cite{lin2020infogan,eastwood2018framework}, an OpenDR-based renderer for teapot images; and \pythonavatar{} \cite{pythonavatar}, a rule-based generator for avatars. 

\myparatightestn{Data-only access.} Since the assets in simulators (e.g., 3D models in Blender) are often proprietary,
many works release only generated datasets without the simulator code. Examples include the \facesynthetics{} \cite{wood2021fake} and the \digiface{} \cite{bae2023digiface} datasets, both produced with Blender-based renderers for human faces.

In \cref{sec:method_simulator,sec:method_data}, we present designs under simulators with code access and data-only access, respectively.

\subsubsection{\simpe{} with Simulator Access}
\label{sec:method_simulator}

While different simulators have very different programming interfaces, most of them can be abstracted in the same way. 
\revisiontmlr{A simulator generates samples based on a set of input \emph{parameters}. These parameters are \emph{not} part of the generated samples themselves; rather, they act as \emph{control signals} that determine certain properties of the outputs.\footnote{For well-documented simulators, the list of parameters is typically readily available. For example, \pythonavatar{}, used in \cref{sec:exp_simulator}, explicitly documents its parameters in the README. Alternatively, one can adopt the approach in \cref{sec:method_data}, which does not require explicit identification of parameters.}}
\revisiontmlr{For example, face image renderers in \citet{wood2021fake,bae2023digiface} generate images conditioned on parameters such as facial pose, lighting intensity, and the IDs of the underlying 3D face models and hairstyles.}

\revisiontmlr{These parameters can be further categorized into two types, depending on whether their values admit a natural ordering.}
\emph{Numerical parameters} refer to those with meaningful ordering (e.g., 1 is closer to 2 than to 3), whether discrete (e.g., $\in \brc{0,1,2}$) or continuous (e.g., $\in [0,2]$). In contrast, \emph{categorical parameters} are those \emph{without} inherent ordering, or where even a small change in the parameter can cause a large, unrelated change in the output.
\revisiontmlr{In the face renderer example above, facial pose and lighting intensity are numerical parameters, as their magnitudes are meaningful and variations lead to smooth changes in the generated images. In contrast, the IDs of the 3D face models and hairstyles are categorical parameters, since different IDs correspond to distinct entities without any notion of proximity or ordering.}

\revisiontmlr{Under this formulation, a simulator $\simulator$ generates a sample denoted by $\simulatorfun{\cate_1,\ldots,\cate_\numcate,\nume_1,\ldots,\nume_\numnume}$, where $\cate_1,\ldots,\cate_\numcate \in \cateset_1 \times \cdots \times \cateset_\numcate$ are $\numcate$ \emph{categorical} parameters and $\nume_1,\ldots,\nume_\numnume \in \numeset_1 \times \cdots \times \numeset_\numnume$ are $\numnume$ \emph{numerical} parameters.}

For \randomsampleapiname{}, we simply draw each parameter randomly from its corresponding feasible set: %
\begin{align}
    &\randomsampleapiname{} = \simulatorfun{\cate_1,\ldots,\cate_\numcate,\nume_1,\ldots,\nume_\numnume},\label{eq:simulator_random_api}
    &\text{where } \cate_i \sim \uniform{\cateset_i} \text{ and } \nume_i \sim \uniform{\numeset_i}.
\end{align}
Here, $\uniform{S}$ denotes drawing a sample uniformly at random from the set $S$.

For \samplevariationapiname{}, we generate variations by perturbing the input simulator parameters. For numerical parameters $\nume_i$, we simply add noise. However, for categorical parameters $\cate_i$, where no natural ordering exists among feasible values in $\cateset_i$, adding noise is not applicable. Instead, we re-draw the parameter from the entire feasible set $\cateset_i$ with a certain probability. Formally, it is defined as 
\begin{align}
    & \quad\quad\samplevariationapi{\simulatorfun{\cate_1,\ldots,\cate_\numcate,\nume_1,\ldots,\nume_\numnume}} %
    =\simulatorfun{\cate_1',\ldots,\cate_\numcate',\nume_1',\ldots,\nume_\numnume'},\label{eq:simulator_variation_api}\\
    &\text{where } \nume_i'\sim \uniform{\brb{\nume_i-\numevariationdegree,\nume_i+\numevariationdegree} \cap \numeset_i}
    \text{ and } %
    \cate_i'=\begin{cases}
      \uniform{\cateset_i}, & \text{with probability } \catevariationdegree \nonumber\\
      \cate_i, & \text{with probability } 1-\catevariationdegree
    \end{cases}\nonumber.
\end{align}
Here, $\numevariationdegree$ and $\catevariationdegree$ control the degree of variation. At one extreme, when $\numevariationdegree=\infty$ and $\catevariationdegree=1$, \samplevariationapiname{} completely discards the information of the input sample and reduces to \randomsampleapiname{}. Conversely, when $\numevariationdegree=\catevariationdegree=0$, \samplevariationapiname{} outputs the input sample unchanged.

\subsubsection{\simpe{} with Simulator-generated Data}
\label{sec:method_data}

We assume a simulator-generated dataset of $\numsimdata$ samples, $\simdataset = \brc{\simdata_1, \ldots, \simdata_\numsimdata}$. The goal is to select $\numgensamples$ samples to form the DP synthetic dataset $\generatedsampleset$. Although this problem appears simple, it requires careful design. We illustrate this by explaining why two straightforward approaches fail.

\myparatightestn{Baseline 1: Applying \dpvotingfunctionname{} on $\generatedsampleset$.} One immediate solution is to apply \dpvotingfunctionname{} in \pe{} (\cref{alg:voting}) by treating $\simdataset$ as the generated set $S$. In other words, each private sample votes for its nearest neighbor in $\simdataset$, and the final histogram, aggregating all votes, is privatized with Gaussian noise. We then draw samples from $\simdataset$ according to the privatized histogram (i.e., \cref{line:draw} in \cref{alg:main_full}) to obtain $\generatedsampleset$.

However, the size of the simulator-generated dataset (i.e., $\numsimdata$) is typically very large (e.g., 1.2 million in \citet{bae2023digiface}), and the total amount of added Gaussian noise grows with $\numsimdata$. This means that the resulting histogram suffers from a low signal-to-noise ratio, leading to poor fidelity in $\generatedsampleset$.

\myparatightestn{Baseline 2: Applying \dpvotingfunctionname{} on cluster centers of $\generatedsampleset$.} 
To improve the signal-to-noise ratio of the histogram, one solution is to have private samples vote on the cluster centers of $\simdataset$ instead of the raw samples. Specifically, we first cluster the samples in $\simdataset$ into $\numclusters$ clusters with centers $\brc{\clustercenter_1,\ldots,\clustercenter_{\numclusters}}$ and have private samples vote on these centers rather than individual samples in $\simdataset$.\footnote{Note that voting in \pe{} \citep{lin2023differentially} is conducted in the embedding space. Here, $\clustercenter_i$s represent cluster centers in the embedding space, and each private sample uses its image embedding to find the nearest cluster center.}
Since the number of bins in the histogram decreases from $\numsimdata$ to $\numclusters$, the signal-to-noise ratio improves. Following the approach of the previous baseline, we then draw $\numgensamples$ cluster centers (with replacement) based on the histogram and randomly select a sample from each chosen cluster to construct the final $\generatedsampleset$.

However, when the total number of samples $\numsimdata$ is large, each cluster may contain a diverse set of samples, including those both close to and far from the private dataset. While DP voting on clusters improves the accuracy of the DP histogram and helps select better clusters, there remains a risk of drawing unsuitable samples from the chosen clusters.

\myparatightestn{Our approach.} Our key insight is that the unavoidable trade-off between DP histogram accuracy and selection precision (clusters vs. individual samples) stems from forcing private samples to consider all of $\simdataset$—either directly (baseline 1) or via cluster centers (baseline 2). But this is not necessary: if a sample %
is far from the private data, its nearest neighbors in $\simdataset$ are likely far too (see \cref{app:justification_data_pe} for experiments). %
Thus, we can avoid wasting privacy budget on evaluating such samples.

The iterative selection and refinement process in \pe{} naturally aligns with this idea. For each sample $\simdata_i$, we define its nearest neighbors in $\simdataset$ as $\nnsample^i_1,\ldots,\nnsample^i_\numsimdata$, sorted by closeness, where $\nnsample^i_1 = \simdata_i$ is the closest.
We define \randomsampleapiname{} as drawing a random sample from $\simdataset$:
\begin{align*}
\randomsampleapiname{} \sim \uniform{\simdataset}.
\end{align*}
Since we draw only $\numgensamples$ samples (instead of all $\numsimdata$) from \randomsampleapiname{}, the DP histogram has a higher signal-to-noise ratio. In the following steps (\cref{line:voting,line:normalization,line:draw} in \cref{alg:main_full}), we discard samples far from the private data indicated by the DP histogram, and apply \samplevariationapiname{} only to the remaining as follows:
\begin{align*}
\samplevariationapi{\simdata_i} = \uniform{\brc{\nnsample^i_1,\ldots,\nnsample^i_\nnvariationdegree}},
\end{align*}
thus avoiding consideration of nearest neighbors of the removed samples (unless they are also nearest neighbors of retained samples).
Similar to $\numevariationdegree$ and $\catevariationdegree$ in \cref{sec:method_simulator}, $\nnvariationdegree$ controls the degree of variation. At one extreme, when $\nnvariationdegree = \numsimdata$, \samplevariationapiname{} disregards the input sample and reduces to \randomsampleapiname{}. At the other extreme, when $\nnvariationdegree = 1$, \samplevariationapiname{} returns the input sample unchanged. 

We verify the effectiveness of these designs in \cref{app:exp_data_selection}.

\myparatightestn{Broader applications.}
While our main experiments (\cref{sec:exp}) focus on simulator-generated data, the proposed algorithm also applies to any public datasets, as shown in \cref{app:cifar10}.

\subsection{\simpe{} with both Simulators and  Models}
\label{sec:method_simulator_and_model}

As discussed in \cref{sec:preliminaries}, simulators and foundation models complement each other across different data domains. Moreover, even within a single domain, they excel in different aspects. For example, computer graphics-based face image %
generators \cite{bae2023digiface,wood2021fake} allow controlled diversity in race, lighting, and makeup while mitigating potential biases in foundation models. However, the generated faces may appear less realistic than those from %
SoTA foundation models. Thus, combining the strengths of both methods for DP data synthesis is highly appealing. 

We find that \simpe{} naturally supports this integration, as  \randomsampleapiname{} and \samplevariationapiname{} work the same for both foundation models and simulators. Specifically, we explore a simple strategy: using simulators in the early \pe{} iterations to generate diverse seed samples, then switching to foundation models in later iterations to refine details and enhance realism. As shown in \cref{sec:exp}, this approach outperforms using either simulators or foundation models alone. We discuss more alternative approaches in \cref{app:simulator_and_model_order}.

\subsection{Enabling New Applications with \simpe{}}
\label{sec:method_new_application}

Prior work (\cref{sec:preliminaries}) typically assumes that the private observations and the target data to be generated lie in the same domain (e.g., generating face images from private ones). However, in many applications, the observed data is \emph{not} the ideal generation target. Instead, the goal is to generate \emph{latent control signals} whose execution in a system reproduces the observed behavior. For example, given recordings of human behavior or gameplay, one may wish to generate robot or player control signals rather than the observations themselves. Such unobserved control signals are more useful for downstream tasks such as training robot models and game-playing agents \cite{argall2009survey,hester2018deep}.

Existing training-based methods (\cref{sec:preliminaries}), which assume that the private and generated data lie in the same space, are \emph{not} applicable to this setting. A natural extension—feeding generated samples into the system (e.g., a robot or game engine) and defining a loss on the resulting outputs—fails because the system is generally non-differentiable.

In contrast, \simpe{} naturally supports these applications by optimizing over simulator parameters (e.g., control signals) that are decoupled from the private observations in the simulator’s output space. By treating the system itself as a simulator, we can apply the algorithm in \cref{sec:method_simulator} to these settings. We demonstrate this in \cref{sec:exp_new_application}.

\section{Evaluation Setup}
\label{sec:exp_setup}

We consider two settings: (1) the traditional case where the private and target generated data lie in the same domain; and (2) a new setting in which they lie in different domains, where prior methods are not directly applicable (\cref{sec:method_new_application}).

\subsection{The Traditional Application}

We use widely studied \emph{DP image generation} benchmarks.

\subsubsection{Datasets and Simulators}
\label{sec:datasets_and_simulators}

\myparatightestn{Datasets.} Following prior work \cite{dpimagebench}, we use three private datasets: \textbf{(1) \mnist{}} \cite{lecun1998mnist}, where the image class labels are digits `0'-`9', \textbf{(2) \celeba{}} \cite{liu2015faceattributes}, where the image class labels are male and female, and
\textbf{(3) \cifar{}} \cite{krizhevsky2009learning}, which contains 10 classes of natural images. We aim at \emph{conditional generation} for these datasets (i.e., each generated image is associated with the class label). Due to space constraints, \cifar{} experiments are deferred to \cref{app:cifar10}.

\myparatightestn{Simulators.}
To showcase the broad applicability of \simpe{}, we use four diverse simulators. %

\noindent\textbf{(1) Text rendering program.} Generating images with readable text using foundation models is a known challenge \cite{betker2023improving}. Simulators can address this gap, as generating text images through computer programs is straightforward. To illustrate this, we use Python PIL library as the text rendering program and treat \mnist{} as the private dataset. 
There are two categorical and three numerical parameters in total (details in \cref{app:mnist_exp_details}).
We set the feasible sets of these parameters to be large enough so that the random samples differ significantly from \mnist{} (see \cref{fig:mnist_simulator}).

\noindent\textbf{(2) Computer graphics-based renderer for face images.} Computer graphics-based rendering is widely used in real-world applications such as game development, cartoons, and movie production. This experiment aims to assess whether these advanced techniques can be adapted for DP synthetic image generation via \simpe{}. We use \celeba{} as the private dataset and a Blender-based face image renderer from \citet{bae2023digiface} as the API. Since the source code for their renderer is not publicly available, we apply our data-based algorithm from \cref{sec:method_data} on their released dataset of 1.2 million face images.
Note that this renderer may not necessarily represent the SoTA. 
As visualized in \cref{fig:celeba_simulator}, the generated faces exhibit various unnatural artifacts and appear less realistic than images produced by SoTA generative models (e.g., \citet{rombach2022high}). Therefore, this experiment serves as a preliminary study, and \emph{the results could potentially improve with more advanced rendering techniques.}

\noindent\textbf{(3) Rule-based avatar generator.} We further investigate whether \simpe{} remains effective \emph{when the simulator's data significantly differs from the private dataset.} We use \celeba{} as the private dataset and a rule-based avatar generator \cite{pythonavatar} as the API. 
This simulator has 16 categorical parameters that control attributes of the avatar, including eyes, noses, background colors, skin colors, etc (details in \cref{app:avatar_exp_details}).
As in \cref{fig:celeba_avatar_simulator}, the generated avatars have a cartoon-like appearance and lack fine-grained details, which contrasts sharply with real human face photos in \celeba{}.

\myparatightestn{(4) Public images.} To demonstrate that \simpe{} (\cref{sec:method_data}) generalizes beyond simulator-generated data, we evaluate it using \imagenet{} as $\simdataset$. %
(Results deferred to \cref{app:cifar10}.)

\myparatightestn{Class label information from simulators.}
In main experiments, we assume the simulator does not provide class label information (denoted as ``\classunavail{}''). In \cref{app:class_label_info_from_simulators}, we further investigate a setting where the simulator provides class labels alongside generated images and find that this improves performance. See \cref{app:class_label_info_from_simulators} for details and results.

\subsubsection{Metrics and Evaluation Pipelines}

We follow the evaluation settings of DPImageBench \cite{dpimagebench}, a recent benchmark for DP image synthesis. Specifically, we use two metrics: \textbf{(1) Fidelity: FID} \cite{heusel2017gans}, which computes the distribution distance between real and generated images in the embedding space; \textbf{(2) Utility: the accuracy of downstream classifiers} trained on the synthetic data.
We carefully design the evaluation process so that the reported accuracy is not inflated due to train-test overlap or DP violations in classifier hyperparameter tuning. See \cref{app:metrics} for the detailed procedure and the rationale behind our choice of metrics.

Following \citet{dpimagebench}, we set DP parameter $\delta=1/(\numprisamples\cdot\log \numprisamples)$, where $\numprisamples$ is the number of samples in the private dataset, and $\epsilon=1$ or 10.

\subsubsection{Baselines}

We compare \simpe{} with 12 SoTA DP image synthesizers reported in \citet{dpimagebench}, including DP-MERF \cite{dp-merf}, DP-NTK \cite{dp-ntk}, DP-Kernel \cite{dp-kernel}, GS-WGAN \cite{gs-wgan}, DP-GAN \cite{dpgan}, DPDM \cite{dpdm}, PDP-Diffusion \cite{dp-diffusion}, DP-LDM \cite{dpldm}, DP-LoRA \cite{dplora}, PrivImage \cite{li2023privimage}, and \pe{} with foundation models \cite{lin2023differentially}. Except for \pe{}, all other baselines require model training.
When experimenting with simulator-generated data, we additionally compare \simpe{} against the two baselines introduced in \cref{sec:method_data} (see \cref{app:exp_data_selection}).

\textbf{Note:} %
Different methods rely on different prior knowledge. For instance, many baselines use pre-trained models or public data (see Tab. 2 of \citet{dpimagebench}), whereas \simpe{} does not. Instead, \simpe{} leverages simulators, which are not used by others. \textbf{As such, \simpe{} represents a new evaluation setting, and baseline results serve primarily to contextualize this new paradigm and inspire future work, rather than an apple-to-apple comparison.}

\subsection{The New Application}

\myparatightestn{Environment.}
We use the popular Gymnasium library\footnote{\url{https://github.com/Farama-Foundation/Gymnasium}, previously known as the Gym library \cite{gym} by OpenAI.} \cite{towers2024gymnasium} for our experiments. Gymnasium provides a wide range of simulated environments for robotics, control, physics, and games, originally designed for developing and benchmarking reinforcement learning algorithms. These environments align well with the motivating scenarios in \cref{sec:method_new_application}.

Specifically, we use the \cartpole{} environment in the Gymnasium library as the example. In this environment, a robot cart balances a pole placed vertically on top of it. At each timestep, the \emph{observation} consists of four quantities: cart position, cart velocity, pole angle, and pole angular velocity. A \emph{control signal} moves the cart left or right to keep the pole balanced. The episode terminates when the pole angle exceeds $\pm 12^\circ$ or the cart moves outside the display. The task is considered successful if the pole remains balanced for 500 frames. See \cref{sec:app_cartpole_settings} for more implementation details.

\revisiontmlr{\myparatightestn{\simpe{} adoption.} We follow the approach described in \cref{sec:method_new_application}. In particular, we use four numerical parameters that map \emph{observations} to \emph{control signals} via a linear function. See \cref{sec:app_cartpole_settings} for additional implementation details.}

\myparatightestn{Metrics.} We propose two metrics:
\textbf{(1) Utility}: For each generated policy, we run it in the simulator to determine whether it succeeds or fails, compare this outcome to the assigned label, and report the accuracy. This metric directly aligns with the objective of the environment—successfully balancing the pole.
\textbf{(2) Fidelity}: We compute the Wasserstein distance between the distributions of episode lengths (i.e., observation sequence lengths) in the generated and private datasets. This provides a more fine-grained measure than utility alone, as episode lengths can vary significantly even among failed policies.

\section{Results}
\label{sec:exp}

We focus on six research questions:\\
\textbf{(1)} Does \simpe{} improve upon \pe{} (\cref{sec:exp_simpe_better})?\\
\textbf{(2)} Does \simpe{} work with \emph{bad} simulators (\cref{sec:result_simpe_weak_simulator})?\\
\textbf{(3)} Can \simpe{} leverage the strengths of both simulators and foundation models (\cref{sec:result_simpe_both})?\\
\textbf{(4)} How does \simpe{} perform in new application settings where prior methods are inapplicable (\cref{sec:exp_new_application})?\\
\textbf{(5)} How efficient is \simpe{} (\cref{sec:exp_further})?\\
\textbf{(6)} How sensitive is \simpe{} to hyperparameters, and how does it perform without embedding models (\cref{sec:exp_further})?

\subsection{\simpe{} Improves \pe{}}
\label{sec:exp_simpe_better}
\subsubsection{\simpe{} with Simulator Access}
\label{sec:exp_simulator}

We evaluate \simpe{} with a text rendering program on \mnist{}. The results (\cref{tab:acc_and_fid,fig:mnist}) show that:

\begin{figure*}[!t]
\centering
\begin{minipage}{0.32\textwidth}
  \centering
\includegraphics[width=0.6\textwidth]{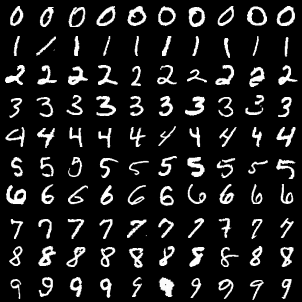}
\subcaption{Real (private) images}\label{fig:mnist_real}
\end{minipage}%
\begin{minipage}{0.32\textwidth}
  \centering
\includegraphics[width=0.6\textwidth]{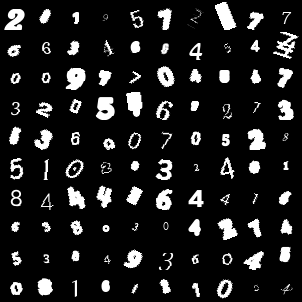}
\subcaption{Simulator-generated images}\label{fig:mnist_simulator}
\end{minipage}%
\begin{minipage}{0.32\textwidth}
  \centering
\includegraphics[width=0.6\textwidth]{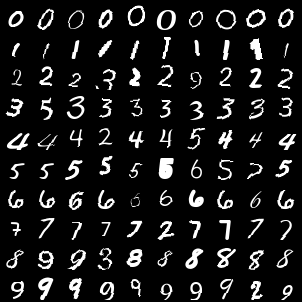}
\subcaption{\simpe{} images ($\epsilon=10$)}\label{fig:mnist_simpe}
\end{minipage}
\caption{The real and generated images on \mnist{} under the ``\classunavail{}'' setting. Each row corresponds to one class.
The simulator generates images that are very different from the real ones and are from the incorrect classes. Starting from these bad images, \simpe{} can effectively guide the generation of the simulator towards high-quality images with correct classes.} \label{fig:mnist}
\end{figure*}
\begin{table}[!t]
\centering

\caption{Accuracy and FID. The best between \pe{} methods is in \textbf{bold}, and the best between all methods is \underline{underlined}. ``Simulator'' refers to samples from the simulator's \randomsampleapiname{}. %
    Results other than \simpe{} and Simulator are taken from \citet{dpimagebench}.}
    \label{tab:acc_and_fid}
\begin{subtable}{0.5\linewidth}
\caption{Accuracy (\%) of downstream classifiers.}\label{tab:classifier}
    \centering
\resizebox{0.8\textwidth}{!}{
    \begin{tabular}{l|cc|cc}
    \toprule
    \multirow{2}{*}{Algorithm} & \multicolumn{2}{c|}{{\mnist{}}} & \multicolumn{2}{c}{{\celeba{}}}\\
    \cline{2-5}
     & $\epsilon = 1$ & $\epsilon = 10$ &  $\epsilon = 1$ & $\epsilon = 10$  \\
    \hline
    DP-MERF & 80.3 & 81.3 & 81.0 & 81.2 \\
    DP-NTK & 50.0 & 91.3 & 61.2 & 64.2 \\
    DP-Kernel & 94.0 & 93.6 & 83.0 & 83.7 \\
    GS-WGAN & 72.4 & 75.3 & 61.4 & 61.5 \\
    DP-GAN & 92.4 & 92.7 & 77.9 & 89.2 \\
    DPDM & 89.2 & 97.7 & 74.5 & 91.8 \\
    PDP-Diffusion & \underline{94.5}  & 97.4 & 89.4 & \underline{94.0} \\
    DP-LDM (SD) & 78.8 & 94.4 & 84.4 & 89.1 \\
    DP-LDM & 44.2 & 95.5 & 85.8 & 92.4 \\
    DP-LoRA & 82.2 & 97.1 & 87.0 & 92.0 \\
    PrivImage & 94.0 & \underline{97.8} & \underline{90.8} & 92.0\\
    \bottomrule
    Simulator & \multicolumn{2}{c|}{11.6 ($\epsilon=0$)} & \multicolumn{2}{c}{61.4 ($\epsilon=0$)} \\
    \pe{} & 27.9 & 32.7 & 70.5 & 74.2 \\
    \simpe{} (ours) & \textbf{89.1} & \textbf{93.6} & \textbf{80.0} & \textbf{82.5} \\
    \bottomrule
\end{tabular}}
\end{subtable}%
\begin{subtable}{0.5\linewidth}
\caption{FID of synthetic images.}\label{tab:fid}
    \centering
\resizebox{0.8\textwidth}{!}{
    \begin{tabular}{l|cc|cc}
    \toprule
    \multirow{2}{*}{Algorithm} & \multicolumn{2}{c|}{{\mnist{}}} & \multicolumn{2}{c}{{\celeba{}}}\\
    \cline{2-5}
     & $\epsilon = 1$ & $\epsilon = 10$ &  $\epsilon = 1$ & $\epsilon = 10$  \\
    \hline
    DP-MERF & 113.7 & 106.3 & 176.3 & 147.9 \\
    DP-NTK & 382.1 & 69.2 & 350.4 & 227.8 \\
    DP-Kernel & 33.7 & 38.9 & 140.3 & 128.8 \\
    GS-WGAN & 57.0 & 47.7 & 611.8 & 290.0 \\
    DP-GAN & 82.3 & 30.3 & 112.5 & 31.7 \\
    DPDM & 36.1 & 4.4 & 153.99 & 28.8 \\
    PDP-Diffusion & 8.9 & 3.8 & 17.1 & \underline{8.1} \\
    DP-LDM (SD) & 31.9 & 18.7 & 46.2 & 24.1 \\
    DP-LDM & 155.2 & 99.1 & 124.1 & 40.4 \\
    DP-LoRA & 112.8 & 95.4 & 53.3 & 32.2 \\
    PrivImage & \underline{7.6} & \underline{2.3} & \underline{11.4} & 11.3\\
    \bottomrule
    Simulator & \multicolumn{2}{c|}{86.2 ($\epsilon=0$)} & \multicolumn{2}{c}{37.2 ($\epsilon=0$)} \\
    \pe{} & 48.8 & 45.3 & \textbf{23.4} & 22.0 \\
    \simpe{} (ours) & \textbf{20.7} & \textbf{9.4} & 24.7 & \textbf{20.8} \\
    \bottomrule
\end{tabular}}

\end{subtable}
\end{table}
\begin{table}[!t]
    \centering
    \caption{
    Accuracy (\%) of classifiers trained on synthetic images and FID of synthetic images on \celeba{}. The best results are highlighted in bold. Using a combination of both (weak) simulators and foundation models outperforms using either one alone.}
    \label{tab:celeba_avatar}
    \vspace{-0.2cm}
    \resizebox{0.6\linewidth}{!}{
    \begin{tabular}{l|cc|cc}
    \toprule
    \multirow{2}{*}{Algorithm} & \multicolumn{2}{c|}{{FID $\downarrow$}} & \multicolumn{2}{c}{{Classification Acc. $\uparrow$}}\\
    \cline{2-5}
     & $\epsilon = 1$ & $\epsilon = 10$ &  $\epsilon = 1$ & $\epsilon = 10$  \\
     \hline
    \pe{} with foundation models & 23.4 & 22.0 & 70.5 & 74.2 \\
    \pe{} with weak simulators (i.e., \simpe{}) & 101.4 & 99.5 & 62.6 & 63.2 \\
    \pe{} with both & \textbf{15.0} & \textbf{11.9} & \textbf{72.7} & \textbf{78.1} \\
    \bottomrule
\end{tabular}}
\end{table}

\myparatightestn{\simpe{} effectively guides the simulator to generate high-quality samples.}
As shown in \cref{fig:mnist_simulator}, without any information from the private data or guidance from \simpe{}, the simulator initially produces poor-quality images with incorrect digit sizes, rotations, and stroke widths. These low-quality samples serve as the starting point for \simpe{} (via \randomsampleapiname{}). Through iterative refinement and private data voting, \simpe{} gradually optimizes the simulator parameters, ultimately generating high-quality \mnist{} samples, as illustrated in \cref{fig:mnist_simpe}.

Quantitative results in \cref{tab:acc_and_fid} further support this. Without private data guidance, the simulator often generates digits from incorrect classes, yielding a classifier accuracy of just 11.6\%--near random guess. In contrast, \simpe{} boosts accuracy to around 90\%. FID scores also confirm that \simpe{} produces images more similar to real data.

\myparatightestn{\simpe{} can improve the performance of \pe{} by a large margin.} 
The \pe{} baseline \cite{lin2023differentially} uses a diffusion model pre-trained on \imagenet{}, which primarily contains natural object images (e.g., plants, animals, cars). Since \mnist{} differs significantly from such data, \pe{}, as a training-free method, struggles to generate meaningful \mnist{}-like images. Most \pe{}-generated images lack recognizable digits (see \citet{dpimagebench}), resulting in a classification accuracy of only $\sim 30\%$ (\cref{tab:classifier}). By leveraging a simulator better suited for this domain, \simpe{} achieves much better results, tripling the classification accuracy and reducing the FID by 80\% at $\epsilon=10$.

\myparatightestn{\simpe{} achieves competitive results among SoTA methods.}
When the foundation model or public data differs significantly from the private data, training-based baselines can still adapt the model to the private data distribution by updating its weights, whereas \pe{} cannot. This limitation accounts for the substantial performance gap between \pe{} and other methods. Specifically, \pe{} records the lowest classification accuracy among all 12 methods (\cref{tab:classifier}). By leveraging domain-specific simulators, \simpe{} substantially narrows this gap, achieving classification accuracy within 5.4\% and 4.2\% of the best-performing method for $\epsilon=1$ and $\epsilon=10$, respectively.

\subsubsection{\simpe{} with Simulator-generated Data}
\label{sec:result_simpe_data}
\begin{figure*}[!t]
\centering

\vspace{-0.cm}
\begin{minipage}{0.32\textwidth}
  \centering
\includegraphics[width=0.95\textwidth]{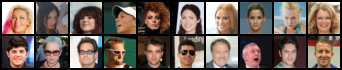}
\subcaption{Real (private) images}\label{fig:celeba_real}
\end{minipage}%
\begin{minipage}{0.32\textwidth}
  \centering
\includegraphics[width=0.95\textwidth]{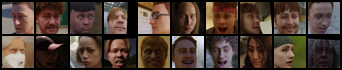}
\subcaption{Simulator-generated images}\label{fig:celeba_simulator}
\end{minipage}%
\begin{minipage}{0.32\textwidth}
  \centering
\includegraphics[width=0.95\textwidth]{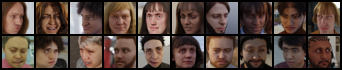}
\subcaption{\simpe{} images ($\epsilon=10$)}\label{fig:celeba_simpe}
\end{minipage}
\vspace{-0.2cm}
\caption{The real and generated images on \celeba{}. The top rows correspond to the ``female'' class, and the bottom rows correspond to the ``male'' class. 
The simulator generates images with incorrect classes. However, starting from these misclassified images, \simpe{} effectively selects those that better match the correct class.} \label{fig:celeba}
\vspace{-0.6cm}
\end{figure*}

We evaluate \simpe{} on \celeba{} with a computer graphics-based renderer. The results (\cref{tab:acc_and_fid,fig:celeba}) show that:

\myparatightestn{\simpe{} selects samples that better match target classes.}
Without private data, the simulator naturally generates images with incorrect labels (\cref{fig:celeba_simulator}). As a result, a gender classifier trained on this data achieves at most 61.4\% accuracy--the majority class (female) rate. \simpe{} iteratively refines selection from this noisy pool, ultimately choosing samples more aligned with target classes (\cref{fig:celeba_simpe}), boosting accuracy by up to 21.1\% (\cref{tab:classifier}).

\myparatightestn{\simpe{} maintains the strong data quality of \pe{}.}
\simpe{} improves \pe{}'s classification accuracy by a large margin (\cref{tab:classifier}) and maintains the good FID (\cref{tab:fid}). Unlike in \mnist{} (\cref{sec:exp_simulator}) where \simpe{} brought large gains, the modest FID improvement on \celeba{} stems from two factors. First, \pe{} with foundation models already ranks 3rd in FID, leaving little room to improve. Second, \simpe{} here only selects from a fixed simulator-generated dataset. As seen in \cref{fig:celeba_simulator}, these images differ from real \celeba{} (e.g., having larger faces), and it is impossible for \simpe{} to %
correct such discrepancies. %
Having access to simulator code, as in \cref{sec:exp_simulator}, could help alleviate such errors by parameter adjustments. A hybrid approach combining foundation models and simulators to be shown in \cref{sec:result_simpe_both} may also help.

\subsection{\simpe{} with Bad Simulators}
\label{sec:result_simpe_weak_simulator}

Here, we examine \simpe{} under bad simulators. As in the previous section, we use \celeba{} as the private dataset, but replace the simulator with a rule-based cartoon avatar generator \cite{pythonavatar}. As shown in \cref{fig:celeba_avatar_simulator}, the generated avatars differ significantly from real %
images.

\myparatightestn{\simpe{} with bad simulators still learns useful features.} From \cref{tab:celeba_avatar}, we observe that downstream classifiers trained on \simpe{} with bad simulators achieve poor classification accuracy. However, two interesting results emerge: \textbf{(1)} Despite the significant difference between avatars and real face images, \simpe{} still captures certain characteristics of the two classes correctly. Specifically, \simpe{} tends to generate faces with long hair for the female class and short hair for the male class (\cref{fig:celeba_avatar_simpe}). \textbf{(2)} Although the FID of \simpe{} is quite poor (\cref{tab:celeba_avatar}), they still outperform many baselines (\cref{tab:fid}). This can be explained by the fact that, as shown in \citet{dpimagebench}, high DP noise makes the training of many baselines unstable. %
This results in images with noisy patterns, non-face images, or significant mode collapse, particularly for DP-NTK, DP-Kernel, and GS-WGAN. In contrast, \simpe{} is training-free, and thus it avoids these issues. See \cref{app:weak_simulator} for more %
analysis.

\subsection{\simpe{} Can Leverage both Simulators and %
Models}
\label{sec:result_simpe_both}

Next, we explore the feasibility of using \pe{} with both foundation models and the bad avatar simulator (\cref{sec:method_simulator_and_model}). 

\myparatightestn{\simpe{} benefits from utilizing simulators and foundation models together.} We observe (\cref{tab:celeba_avatar}) that using both simulators and foundation models yields the best results. This result is intuitive: the foundation model, pre-trained on the diverse \imagenet{} dataset, has a low probability of generating faces via \randomsampleapiname{}. While avatars differ from \celeba{}, they retain the correct image layout, such as facial boundaries, eyes, and nose. Using these avatars as seed samples for variation allows the foundation model to focus on images closer to real faces rather than random patterns.

Unlike other SoTA methods tied to a specific data synthesizer, this result suggests that \pe{} is a promising framework that can combine the strengths of multiple data synthesizers.

\subsection{New Applications Enabled by \simpe{}}
\label{sec:exp_new_application}

\begin{figure*}[!t]
\centering

\vspace{-0.cm}
\begin{minipage}{0.4\textwidth}
  \centering
\includegraphics[width=0.7\textwidth]{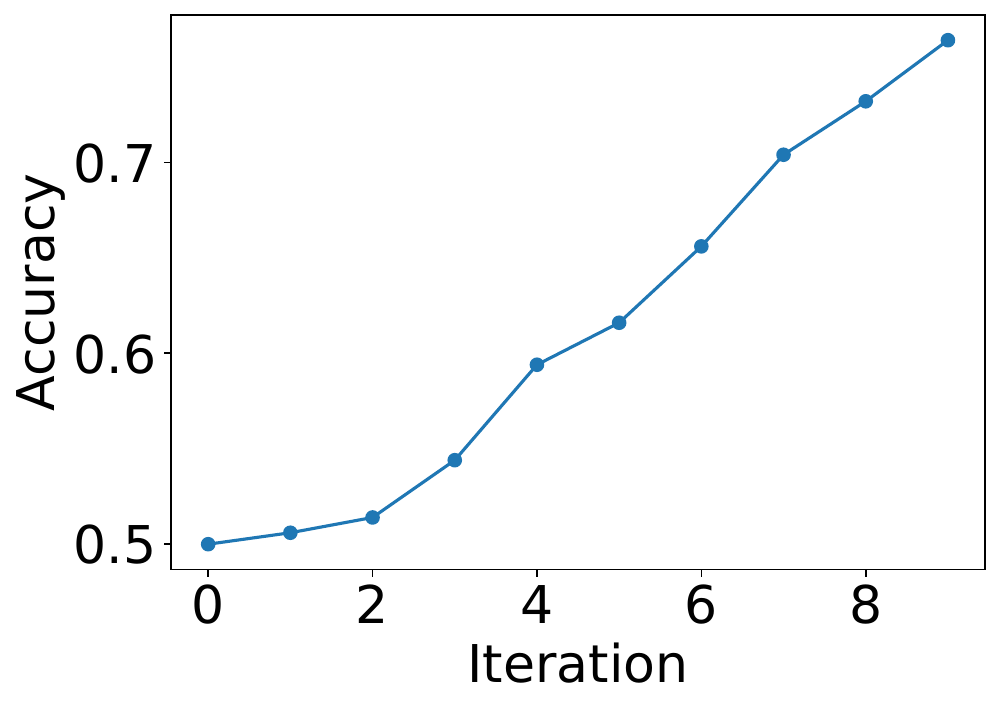}
\subcaption{Accuracy vs. \simpe{} iteration}
\end{minipage}%
\begin{minipage}{0.4\textwidth}
  \centering
\includegraphics[width=0.7\textwidth]{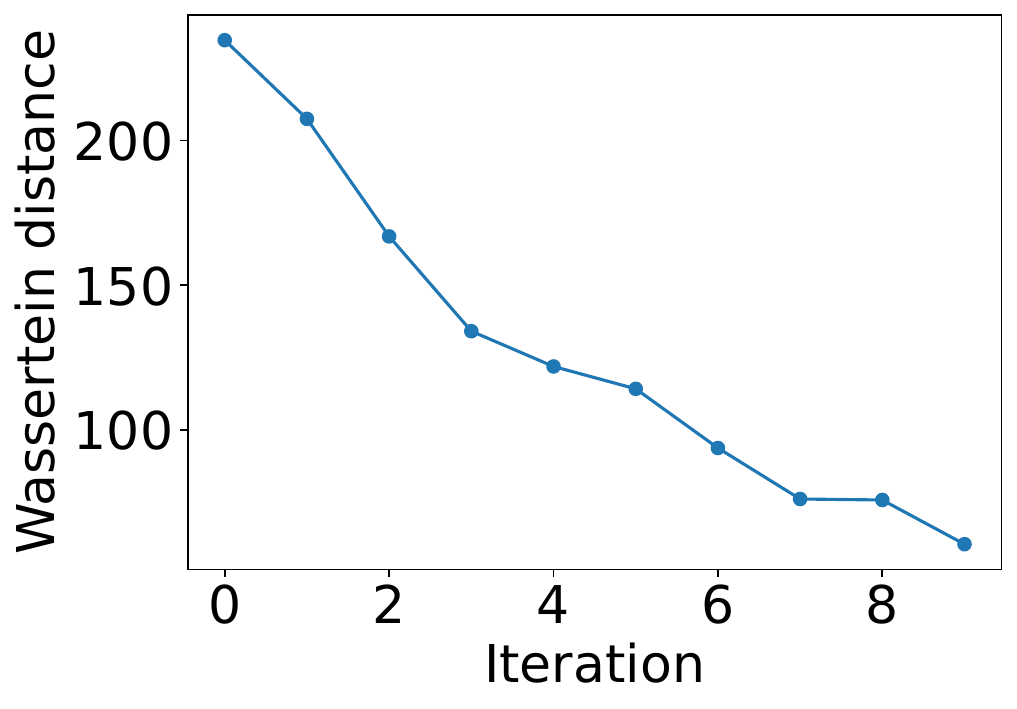}
\subcaption{Wassertein distance vs. \simpe{} iteration}
\end{minipage}%
\caption{The quality of \simpe{}'s generated data improves across iterations in the \cartpole{} environment ($\epsilon=1$). } \label{fig:new_application_iteration}
\vspace{-0.cm}
\end{figure*}
\begin{figure*}[!t]
\centering

\vspace{-0.cm}
\begin{minipage}{0.4\textwidth}
  \centering
\includegraphics[width=0.7\textwidth]{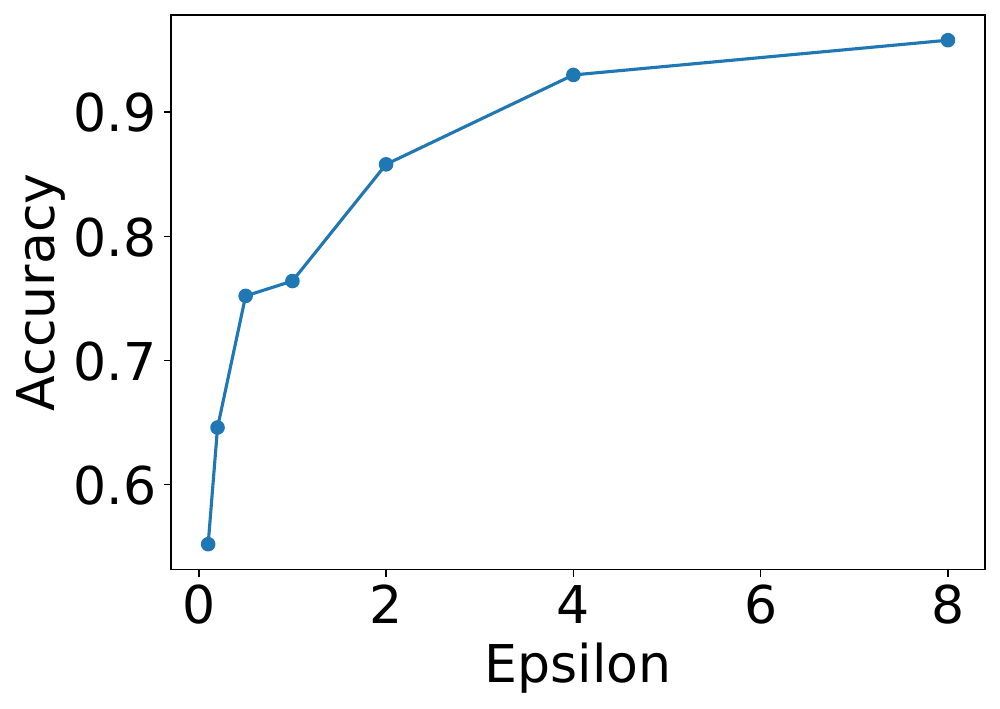}
\subcaption{Accuracy vs. epsilon}
\end{minipage}%
\begin{minipage}{0.4\textwidth}
  \centering
\includegraphics[width=0.7\textwidth]{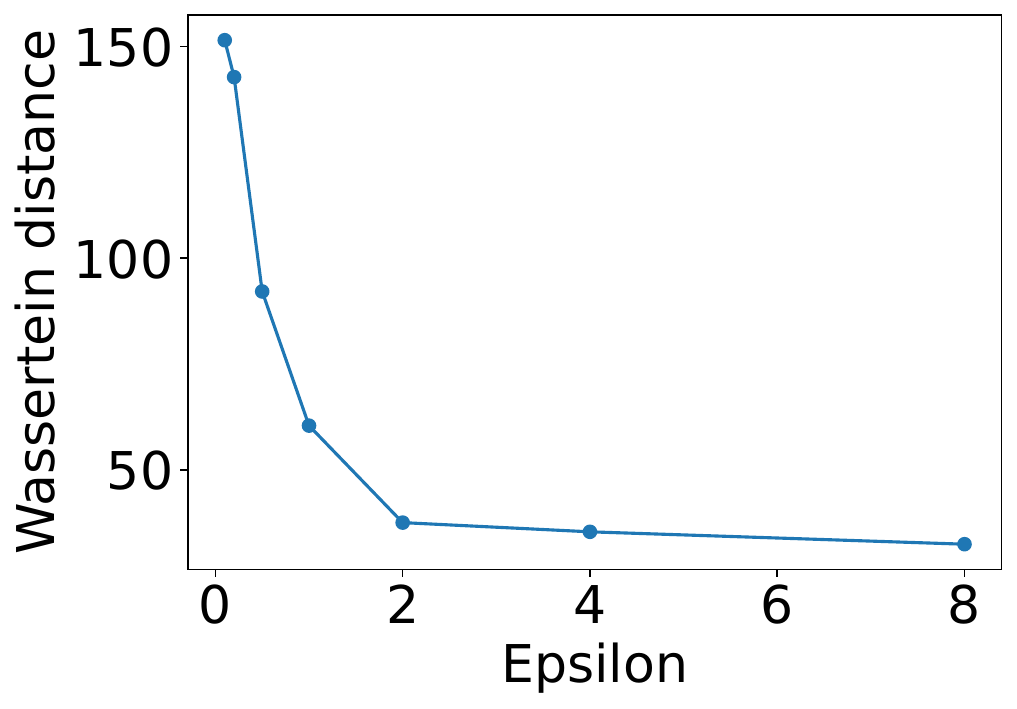}
\subcaption{Wassertein distance vs. epsilon}
\end{minipage}%
\caption{The quality of \simpe{}'s generated data improves with more privacy budget in the \cartpole{} environment. } \label{fig:new_application_epsilon}
\vspace{-0.3cm}
\end{figure*}

Since prior methods are not directly applicable (\cref{sec:method_new_application}), we focus on evaluating the properties of \simpe{}.

\myparatightestn{\simpe{}'s data quality improves across iterations.}
\cref{fig:new_application_iteration} show that both utility and fidelity metrics steadily improve over time. More specifically: (1) since random policies fail roughly 99\% of the time, the initial accuracy is near 50\%. \simpe{} progressively discovers policies that match the success/failure patterns of the private data, ultimately reaching 76\% accuracy. (2) The Wasserstein distance also decreases by roughly $4\times$ over the iterations. These results demonstrate that \simpe{} is effective even in this new application.

\myparatightestn{\simpe{}'s data quality improves as the privacy budget increases.} We vary the privacy budget $\epsilon$ from 0.2 to 8.  \cref{fig:new_application_epsilon} shows a clear trend: performance improves as the privacy budget increases. With $\epsilon=8$, the accuracy reaches 96\%, a very strong result. This confirms that \simpe{} can leverage larger privacy budgets effectively when they are available.

\vspace{-0.1cm}
\subsection{Further Analyses and Discussions} %
\label{sec:exp_further}
\vspace{-0.1cm}

\myparatightestn{\simpe{} converges quickly.} Like \pe{}, \simpe{} achieves good results in a few (e.g., 6) iterations (\cref{fig:metrics_vs_iterations}).

\myparatightestn{\simpe{} is efficient.} Since simulators could be cheaper to run than foundation models, \simpe{} could be much more efficient than \pe{}. For instance, on \mnist{}, each \pe{}’s API call takes over \emph{2400 GPU seconds}, whereas \simpe{} takes less than \emph{30 CPU seconds}--an 80x speedup, not to mention the lower cost of CPU than GPU. (More results in \cref{app:efficiency}.)

\myparatightestn{Embedding model.} 
Computing the sample distance in \cref{line:distance} is performed in an embedding space, following \citet{lin2023differentially}. The \samplevariationapiname{} in \cref{sec:method_data} also retrieves nearest samples in this embedding space. We clarify the following points:
\textbf{(1) Experimental fairness.}
For image experiments, we use embeddings from a pre-trained Inception model, assuming its training data (\imagenet{}) is public and independent of the private dataset. This setup is consistent across methods: \pe{} uses the same ImageNet-pretrained Inception embedding, and all training-based baselines also treat \imagenet{} as public data and rely on it similarly, typically via pretraining \cite{dpimagebench}.
\textbf{(2) Reasonableness of the assumption.}
Sim-PE assumes access to a public embedding model, which aligns with standard assumptions in DP fine-tuning, where large pretrained models are treated as public and non-private. In practice, such pretraining often relies on massive datasets, sometimes including potentially sensitive corpora (e.g., LAION \cite{schuhmann2021laion} in \citet{dpldm}). In contrast, Sim-PE only depends on an Inception embedding trained on ImageNet, making this assumption comparatively weaker.
\textbf{(3) Applicability without pretrained embeddings.}
\simpe{} can also operate without pretrained embeddings in domains lacking public data or strong representation models. For example, in our \cartpole{} experiments (\cref{sec:exp_new_application}), we directly use raw samples as embeddings (\cref{sec:app_cartpole_settings}), demonstrating that \simpe{} could remain effective even without any pretrained embedding.

\myparatightestn{Hyperparameter sensitivity.} We conduct ablation studies in \cref{app:ablation}, including stress tests of \simpe{} under unreasonable parameter settings. Overall, \simpe{} is robust to simulator parameter ranges, its hyperparameters ($\numevariationdegree, \catevariationdegree, \nnvariationdegree$), and the timing of switching from simulators to foundation models, unless these settings are made excessively poor. We also study the effect of sample size and distribution alignment.

\vspace{-0.2cm}
\section{Limitations and Future Work}
\label{sec:discussion}
\vspace{-0.2cm}

In this paper, we demonstrate the potential of the \pe{} framework for utilizing simulators in DP data synthesis. While the results are promising, important questions remain open:
\begin{packeditemize}
    \item 
    Since \simpe{} (in \cref{sec:method_data}) can be applied to any public dataset (\cref{app:cifar10}), it could support applications such as pre-training data selection for private fine-tuning~\cite{yu2023selective,li2023privimage}.
    \item 
    In domains like networking and systems, simulators are more common than foundation models, and \simpe{} could offer even greater potential. We provide preliminary evaluations in \cref{app:network}.
    \item 
    \simpe{} for image synthesis is still outperformed by the best baseline. Exploring better ways to combine simulators and foundation models could further push the limits of the \pe{} framework.
    \item 
    It remains an open question how to effectively apply \pe{} in domains where suitable foundation models, simulators, and public datasets are unavailable. \Cref{app:network} provides initial evidence that, in some domains, \simpe{} can still be effective using only simple random-number generators as its APIs, without relying on sophisticated simulators: the \randomsampleapiname{} generates random numbers, while the \samplevariationapiname{} adds perturbations.
 
\end{packeditemize}
\section*{Acknowledgement}

The authors would like to thank the anonymous reviewers for their valuable feedback and suggestions. 

$^\dagger$ This paper is the full version of our previous workshop papers \cite{lin2025differentiallyworkshop1,lin2025differentiallyworkshop2}.

\bibliography{main}

\begin{thebibliography}{89}
\providecommand{\natexlab}[1]{#1}
\providecommand{\url}[1]{\texttt{#1}}
\expandafter\ifx\csname urlstyle\endcsname\relax
  \providecommand{\doi}[1]{doi: #1}\else
  \providecommand{\doi}{doi: \begingroup \urlstyle{rm}\Url}\fi

\bibitem[Afonja et~al.(2024)Afonja, Sim, Lin, Inan, and Yekhanin]{microsoft_pe}
Gbola Afonja, Robert Sim, Zinan Lin, Huseyin~Atahan Inan, and Sergey Yekhanin.
\newblock The crossroads of innovation and privacy: Private synthetic data for
  generative ai.
\newblock
  \url{https://www.microsoft.com/en-us/research/blog/the-crossroads-of-innovation-and-privacy-private-synthetic-data-for-generative-ai/},
  2024.

\bibitem[Apple(2025)]{apple_pe}
Apple.
\newblock Understanding aggregate trends for apple intelligence using
  differential privacy.
\newblock
  \url{https://machinelearning.apple.com/research/differential-privacy-aggregate-trends},
  2025.

\bibitem[Argall et~al.(2009)Argall, Chernova, Veloso, and
  Browning]{argall2009survey}
Brenna~D Argall, Sonia Chernova, Manuela Veloso, and Brett Browning.
\newblock A survey of robot learning from demonstration.
\newblock \emph{Robotics and autonomous systems}, 57\penalty0 (5):\penalty0
  469--483, 2009.

\bibitem[Authors(2024)]{Genesis}
Genesis Authors.
\newblock Genesis: A universal and generative physics engine for robotics and
  beyond, December 2024.
\newblock URL \url{https://github.com/Genesis-Embodied-AI/Genesis}.

\bibitem[Bae et~al.(2023)Bae, de~La~Gorce, Baltru{\v{s}}aitis, Hewitt, Chen,
  Valentin, Cipolla, and Shen]{bae2023digiface}
Gwangbin Bae, Martin de~La~Gorce, Tadas Baltru{\v{s}}aitis, Charlie Hewitt,
  Dong Chen, Julien Valentin, Roberto Cipolla, and Jingjing Shen.
\newblock Digiface-1m: 1 million digital face images for face recognition.
\newblock In \emph{Proceedings of the IEEE/CVF Winter Conference on
  Applications of Computer Vision}, pp.\  3526--3535, 2023.

\bibitem[Balle \& Wang(2018)Balle and Wang]{balle2018improving}
Borja Balle and Yu-Xiang Wang.
\newblock Improving the gaussian mechanism for differential privacy: Analytical
  calibration and optimal denoising.
\newblock In \emph{International Conference on Machine Learning}, pp.\
  394--403. PMLR, 2018.

\bibitem[Beaulieu-Jones et~al.(2019)Beaulieu-Jones, Wu, Williams, Lee,
  Bhavnani, Byrd, and Greene]{beaulieu2019privacy}
Brett~K Beaulieu-Jones, Zhiwei~Steven Wu, Chris Williams, Ran Lee, Sanjeev~P
  Bhavnani, James~Brian Byrd, and Casey~S Greene.
\newblock Privacy-preserving generative deep neural networks support clinical
  data sharing.
\newblock \emph{Circulation: Cardiovascular Quality and Outcomes}, 12\penalty0
  (7):\penalty0 e005122, 2019.

\bibitem[Betker et~al.(2023)Betker, Goh, Jing, Brooks, Wang, Li, Ouyang,
  Zhuang, Lee, Guo, et~al.]{betker2023improving}
James Betker, Gabriel Goh, Li~Jing, Tim Brooks, Jianfeng Wang, Linjie Li, Long
  Ouyang, Juntang Zhuang, Joyce Lee, Yufei Guo, et~al.
\newblock Improving image generation with better captions.
\newblock \emph{Computer Science. https://cdn. openai. com/papers/dall-e-3.
  pdf}, 2\penalty0 (3):\penalty0 8, 2023.

\bibitem[Bowen \& Snoke(2019)Bowen and Snoke]{bowen2019comparative}
Claire~McKay Bowen and Joshua Snoke.
\newblock Comparative study of differentially private synthetic data algorithms
  from the nist pscr differential privacy synthetic data challenge.
\newblock \emph{arXiv preprint arXiv:1911.12704}, 2019.

\bibitem[Brockman et~al.(2016)Brockman, Cheung, Pettersson, Schneider,
  Schulman, Tang, and Zaremba]{gym}
Greg Brockman, Vicki Cheung, Ludwig Pettersson, Jonas Schneider, John Schulman,
  Jie Tang, and Wojciech Zaremba.
\newblock Openai gym, 2016.

\bibitem[Cao et~al.(2021)Cao, Bie, Vahdat, Fidler, and Kreis]{cao2021don}
Tianshi Cao, Alex Bie, Arash Vahdat, Sanja Fidler, and Karsten Kreis.
\newblock Don’t generate me: Training differentially private generative
  models with sinkhorn divergence.
\newblock \emph{Advances in Neural Information Processing Systems},
  34:\penalty0 12480--12492, 2021.

\bibitem[Chen et~al.(2020)Chen, Orekondy, and Fritz]{gs-wgan}
Dingfan Chen, Tribhuvanesh Orekondy, and Mario Fritz.
\newblock {GS-WGAN:} {A} gradient-sanitized approach for learning
  differentially private generators.
\newblock In \emph{Advances in Neural Information Processing Systems}, 2020.

\bibitem[Chen et~al.(2022)Chen, Yu, Kao, Pang, and Lu]{chen2022dpgen}
Jia-Wei Chen, Chia-Mu Yu, Ching-Chia Kao, Tzai-Wei Pang, and Chun-Shien Lu.
\newblock Dpgen: Differentially private generative energy-guided network for
  natural image synthesis.
\newblock In \emph{Proceedings of the IEEE/CVF Conference on Computer Vision
  and Pattern Recognition}, pp.\  8387--8396, 2022.

\bibitem[Chien et~al.(2026)Chien, Hu, McKenna, Wu, Xu, and
  Kairouz]{chien2026maple}
Eli Chien, Yuzheng Hu, Ryan McKenna, Shanshan Wu, Zheng Xu, and Peter Kairouz.
\newblock Maple: Metadata augmented private language evolution.
\newblock \emph{arXiv preprint arXiv:2603.19258}, 2026.

\bibitem[Community(2018)]{blender}
Blender~Online Community.
\newblock \emph{Blender - a 3D modelling and rendering package}.
\newblock Blender Foundation, Stichting Blender Foundation, Amsterdam, 2018.
\newblock URL \url{http://www.blender.org}.

\bibitem[Devaranjan et~al.(2020)Devaranjan, Kar, and
  Fidler]{devaranjan2020meta}
Jeevan Devaranjan, Amlan Kar, and Sanja Fidler.
\newblock Meta-sim2: Unsupervised learning of scene structure for synthetic
  data generation.
\newblock In \emph{Computer Vision--ECCV 2020: 16th European Conference,
  Glasgow, UK, August 23--28, 2020, Proceedings, Part XVII 16}, pp.\  715--733.
  Springer, 2020.

\bibitem[Dockhorn et~al.(2022)Dockhorn, Cao, Vahdat, and
  Kreis]{dockhorn2022differentially}
Tim Dockhorn, Tianshi Cao, Arash Vahdat, and Karsten Kreis.
\newblock Differentially private diffusion models.
\newblock \emph{arXiv preprint arXiv:2210.09929}, 2022.

\bibitem[Dockhorn et~al.(2023)Dockhorn, Cao, Vahdat, et~al.]{dpdm}
Tim Dockhorn, Tianshi Cao, Arash Vahdat, et~al.
\newblock Differentially private diffusion models.
\newblock \emph{Transactions on Machine Learning Research}, 2023.
\newblock ISSN 2835-8856.
\newblock URL \url{https://openreview.net/forum?id=ZPpQk7FJXF}.

\bibitem[Dong et~al.(2022)Dong, Roth, and Su]{dong2022gaussian}
Jinshuo Dong, Aaron Roth, and Weijie~J Su.
\newblock Gaussian differential privacy.
\newblock \emph{Journal of the Royal Statistical Society Series B: Statistical
  Methodology}, 84\penalty0 (1):\penalty0 3--37, 2022.

\bibitem[Dwork et~al.(2006)Dwork, McSherry, Nissim, and
  Smith]{dwork2006calibrating}
Cynthia Dwork, Frank McSherry, Kobbi Nissim, and Adam Smith.
\newblock Calibrating noise to sensitivity in private data analysis.
\newblock In \emph{Theory of Cryptography: Third Theory of Cryptography
  Conference, TCC 2006, New York, NY, USA, March 4-7, 2006. Proceedings 3},
  pp.\  265--284. Springer, 2006.

\bibitem[Dwork et~al.(2014)Dwork, Roth, et~al.]{dwork2014algorithmic}
Cynthia Dwork, Aaron Roth, et~al.
\newblock The algorithmic foundations of differential privacy.
\newblock \emph{Foundations and Trends{\textregistered} in Theoretical Computer
  Science}, 9\penalty0 (3--4):\penalty0 211--407, 2014.

\bibitem[Eastwood \& Williams(2018)Eastwood and
  Williams]{eastwood2018framework}
Cian Eastwood and Christopher~KI Williams.
\newblock A framework for the quantitative evaluation of disentangled
  representations.
\newblock In \emph{6th International Conference on Learning Representations},
  2018.

\bibitem[{Epic Games}()]{unreal}
{Epic Games}.
\newblock Unreal engine.
\newblock https://www.unrealengine.com.
\newblock URL \url{https://www.unrealengine.com}.

\bibitem[Escartín(2021)]{pythonavatar}
Ibon Escartín.
\newblock python avatars.
\newblock \url{https://github.com/ibonn/python_avatars}, 2021.

\bibitem[Ghalebikesabi et~al.(2023{\natexlab{a}})Ghalebikesabi, Berrada, Gowal,
  Ktena, Stanforth, Hayes, De, Smith, Wiles, and
  Balle]{ghalebikesabi2023differentially}
Sahra Ghalebikesabi, Leonard Berrada, Sven Gowal, Ira Ktena, Robert Stanforth,
  Jamie Hayes, Soham De, Samuel~L Smith, Olivia Wiles, and Borja Balle.
\newblock Differentially private diffusion models generate useful synthetic
  images.
\newblock \emph{arXiv preprint arXiv:2302.13861}, 2023{\natexlab{a}}.

\bibitem[Ghalebikesabi et~al.(2023{\natexlab{b}})Ghalebikesabi, Berrada, Gowal,
  et~al.]{dp-diffusion}
Sahra Ghalebikesabi, Leonard Berrada, Sven Gowal, et~al.
\newblock Differentially private diffusion models generate useful synthetic
  images.
\newblock \emph{CoRR}, abs/2302.13861, 2023{\natexlab{b}}.

\bibitem[Gong et~al.(2025)Gong, Li, Lin, and Wang]{dpimagebench}
Chen Gong, Kecen Li, Zinan Lin, and Tianhao Wang.
\newblock Dpimagebench: A unified benchmark for differentially private image
  synthesis.
\newblock \emph{arXiv preprint arXiv:2503.14681}, 2025.

\bibitem[Goodfellow et~al.(2020)Goodfellow, Pouget-Abadie, Mirza, Xu,
  Warde-Farley, Ozair, Courville, and Bengio]{goodfellow2020generative}
Ian Goodfellow, Jean Pouget-Abadie, Mehdi Mirza, Bing Xu, David Warde-Farley,
  Sherjil Ozair, Aaron Courville, and Yoshua Bengio.
\newblock Generative adversarial networks.
\newblock \emph{Communications of the ACM}, 63\penalty0 (11):\penalty0
  139--144, 2020.

\bibitem[Google(2022)]{googlefonts}
Google.
\newblock Google fonts.
\newblock \url{https://github.com/google/fonts}, 2022.

\bibitem[Greff et~al.(2022)Greff, Belletti, Beyer, Doersch, Du, Duckworth,
  Fleet, Gnanapragasam, Golemo, Herrmann, Kipf, Kundu, Lagun, Laradji, Liu,
  Meyer, Miao, Nowrouzezahrai, Oztireli, Pot, Radwan, Rebain, Sabour, Sajjadi,
  Sela, Sitzmann, Stone, Sun, Vora, Wang, Wu, Yi, Zhong, and
  Tagliasacchi]{greff2021kubric}
Klaus Greff, Francois Belletti, Lucas Beyer, Carl Doersch, Yilun Du, Daniel
  Duckworth, David~J Fleet, Dan Gnanapragasam, Florian Golemo, Charles
  Herrmann, Thomas Kipf, Abhijit Kundu, Dmitry Lagun, Issam Laradji,
  Hsueh-Ti~(Derek) Liu, Henning Meyer, Yishu Miao, Derek Nowrouzezahrai, Cengiz
  Oztireli, Etienne Pot, Noha Radwan, Daniel Rebain, Sara Sabour, Mehdi S.~M.
  Sajjadi, Matan Sela, Vincent Sitzmann, Austin Stone, Deqing Sun, Suhani Vora,
  Ziyu Wang, Tianhao Wu, Kwang~Moo Yi, Fangcheng Zhong, and Andrea
  Tagliasacchi.
\newblock Kubric: a scalable dataset generator.
\newblock 2022.

\bibitem[Harder et~al.(2021{\natexlab{a}})Harder, Adamczewski, and
  Park]{dp-merf}
Frederik Harder, Kamil Adamczewski, and Mijung Park.
\newblock {DP-MERF:} differentially private mean embeddings with random
  features for practical privacy-preserving data generation.
\newblock In \emph{{AISTATS}}, pp.\  1819--1827, 2021{\natexlab{a}}.

\bibitem[Harder et~al.(2021{\natexlab{b}})Harder, Adamczewski, and
  Park]{harder2021dp}
Frederik Harder, Kamil Adamczewski, and Mijung Park.
\newblock Dp-merf: Differentially private mean embeddings with randomfeatures
  for practical privacy-preserving data generation.
\newblock In \emph{International conference on artificial intelligence and
  statistics}, pp.\  1819--1827. PMLR, 2021{\natexlab{b}}.

\bibitem[Harder et~al.(2023)Harder, Jalali, Sutherland, and
  Park]{harder2022differentially}
Frederik Harder, Milad Jalali, Danica~J Sutherland, and Mijung Park.
\newblock Pre-trained perceptual features improve differentially private image
  generation.
\newblock \emph{Transactions on Machine Learning Research}, 2023.

\bibitem[He et~al.(2022)He, Li, Yu, Zhang, Kulkarni, Lee, Backurs, Yu, and
  Bian]{he2022exploring}
Jiyan He, Xuechen Li, Da~Yu, Huishuai Zhang, Janardhan Kulkarni, Yin~Tat Lee,
  Arturs Backurs, Nenghai Yu, and Jiang Bian.
\newblock Exploring the limits of differentially private deep learning with
  group-wise clipping.
\newblock \emph{arXiv preprint arXiv:2212.01539}, 2022.

\bibitem[He et~al.(2016)He, Zhang, Ren, and Sun]{he2016deep}
Kaiming He, Xiangyu Zhang, Shaoqing Ren, and Jian Sun.
\newblock Deep residual learning for image recognition.
\newblock In \emph{Proceedings of the IEEE conference on computer vision and
  pattern recognition}, pp.\  770--778, 2016.

\bibitem[Hester et~al.(2018)Hester, Vecerik, Pietquin, Lanctot, Schaul, Piot,
  Horgan, Quan, Sendonaris, Osband, et~al.]{hester2018deep}
Todd Hester, Matej Vecerik, Olivier Pietquin, Marc Lanctot, Tom Schaul, Bilal
  Piot, Dan Horgan, John Quan, Andrew Sendonaris, Ian Osband, et~al.
\newblock Deep q-learning from demonstrations.
\newblock In \emph{Proceedings of the AAAI conference on artificial
  intelligence}, volume~32, 2018.

\bibitem[Heusel et~al.(2017)Heusel, Ramsauer, Unterthiner, Nessler, and
  Hochreiter]{heusel2017gans}
Martin Heusel, Hubert Ramsauer, Thomas Unterthiner, Bernhard Nessler, and Sepp
  Hochreiter.
\newblock Gans trained by a two time-scale update rule converge to a local nash
  equilibrium.
\newblock \emph{Advances in neural information processing systems}, 30, 2017.

\bibitem[Hou et~al.(2024)Hou, Shrivastava, Zhan, Conway, Le, Sagar, Fanti, and
  Lazar]{hou2024pre}
Charlie Hou, Akshat Shrivastava, Hongyuan Zhan, Rylan Conway, Trang Le, Adithya
  Sagar, Giulia Fanti, and Daniel Lazar.
\newblock Pre-text: Training language models on private federated data in the
  age of llms.
\newblock \emph{arXiv preprint arXiv:2406.02958}, 2024.

\bibitem[Hou et~al.(2025)Hou, Wang, Zhu, Lazar, and Fanti]{hou2025private}
Charlie Hou, Mei-Yu Wang, Yige Zhu, Daniel Lazar, and Giulia Fanti.
\newblock Private federated learning using preference-optimized synthetic data.
\newblock \emph{arXiv preprint arXiv:2504.16438}, 2025.

\bibitem[Hu et~al.(2024)Hu, Wu, Li, Long, Garrido, Ge, Ding, Forsyth, Li, and
  Song]{hu2024sok}
Yuzheng Hu, Fan Wu, Qinbin Li, Yunhui Long, Gonzalo~Munilla Garrido, Chang Ge,
  Bolin Ding, David Forsyth, Bo~Li, and Dawn Song.
\newblock Sok: Privacy-preserving data synthesis.
\newblock In \emph{2024 IEEE Symposium on Security and Privacy (SP)}, pp.\
  4696--4713. IEEE, 2024.

\bibitem[Humphries et~al.(2026)Humphries, Lin, and Yekhanin]{humphries2026pe}
Thomas Humphries, Zinan Lin, and Sergey Yekhanin.
\newblock Pe-means: Improved differentially private $ k $-means clustering
  through private evolution.
\newblock \emph{arXiv preprint arXiv:2606.00342}, 2026.

\bibitem[Issariyakul et~al.(2009)Issariyakul, Hossain, Issariyakul, and
  Hossain]{issariyakul2009introduction}
Teerawat Issariyakul, Ekram Hossain, Teerawat Issariyakul, and Ekram Hossain.
\newblock \emph{Introduction to network simulator 2 (NS2)}.
\newblock Springer, 2009.

\bibitem[Jiang et~al.(2023)Jiang, Sun, and Yu]{dp-kernel}
Dihong Jiang, Sun Sun, and Yaoliang Yu.
\newblock Functional renyi differential privacy for generative modeling.
\newblock In \emph{Advances in Neural Information Processing Systems}, 2023.

\bibitem[Jordon et~al.(2019)Jordon, Yoon, and Van Der~Schaar]{jordon2019pate}
James Jordon, Jinsung Yoon, and Mihaela Van Der~Schaar.
\newblock {PATE-GAN}: Generating synthetic data with differential privacy
  guarantees.
\newblock In \emph{International conference on learning representations}, 2019.

\bibitem[Kar et~al.(2019)Kar, Prakash, Liu, Cameracci, Yuan, Rusiniak, Acuna,
  Torralba, and Fidler]{kar2019meta}
Amlan Kar, Aayush Prakash, Ming-Yu Liu, Eric Cameracci, Justin Yuan, Matt
  Rusiniak, David Acuna, Antonio Torralba, and Sanja Fidler.
\newblock Meta-sim: Learning to generate synthetic datasets.
\newblock In \emph{Proceedings of the IEEE/CVF International Conference on
  Computer Vision}, pp.\  4551--4560, 2019.

\bibitem[Krizhevsky et~al.(2009)Krizhevsky, Hinton,
  et~al.]{krizhevsky2009learning}
Alex Krizhevsky, Geoffrey Hinton, et~al.
\newblock Learning multiple layers of features from tiny images.
\newblock 2009.

\bibitem[LeCun(1998)]{lecun1998mnist}
Yann LeCun.
\newblock The mnist database of handwritten digits.
\newblock \emph{http://yann. lecun. com/exdb/mnist/}, 1998.

\bibitem[Li et~al.(2024)Li, Gong, Li, et~al.]{li2023privimage}
Kecen Li, Chen Gong, Zhixiang Li, et~al.
\newblock {PrivImage}: Differentially private synthetic image generation using
  diffusion models with {Semantic-Aware} pretraining.
\newblock In \emph{33rd USENIX Security Symposium (USENIX Security 24)}, pp.\
  4837--4854, 2024.
\newblock ISBN 978-1-939133-44-1.

\bibitem[Li et~al.(2021)Li, Tramer, Liang, and Hashimoto]{li2021large}
Xuechen Li, Florian Tramer, Percy Liang, and Tatsunori Hashimoto.
\newblock Large language models can be strong differentially private learners.
\newblock \emph{arXiv preprint arXiv:2110.05679}, 2021.

\bibitem[Lin(2022)]{lin2022data}
Zinan Lin.
\newblock \emph{Data Sharing with Generative Adversarial Networks: From Theory
  to Practice}.
\newblock PhD thesis, Carnegie Mellon University, 2022.

\bibitem[Lin et~al.(2020{\natexlab{a}})Lin, Jain, Wang, Fanti, and
  Sekar]{lin2020using}
Zinan Lin, Alankar Jain, Chen Wang, Giulia Fanti, and Vyas Sekar.
\newblock Using gans for sharing networked time series data: Challenges,
  initial promise, and open questions.
\newblock In \emph{Proceedings of the ACM Internet Measurement Conference},
  pp.\  464--483, 2020{\natexlab{a}}.

\bibitem[Lin et~al.(2020{\natexlab{b}})Lin, Thekumparampil, Fanti, and
  Oh]{lin2020infogan}
Zinan Lin, Kiran Thekumparampil, Giulia Fanti, and Sewoong Oh.
\newblock Infogan-cr and modelcentrality: Self-supervised model training and
  selection for disentangling gans.
\newblock In \emph{international conference on machine learning}, pp.\
  6127--6139. PMLR, 2020{\natexlab{b}}.

\bibitem[Lin et~al.(2023)Lin, Gopi, Kulkarni, Nori, and
  Yekhanin]{lin2023differentially}
Zinan Lin, Sivakanth Gopi, Janardhan Kulkarni, Harsha Nori, and Sergey
  Yekhanin.
\newblock Differentially private synthetic data via foundation model {API}s 1:
  Images.
\newblock In \emph{NeurIPS 2023 Workshop on Synthetic Data Generation with
  Generative AI}, 2023.
\newblock URL \url{https://openreview.net/forum?id=7GbfIEvoS8}.

\bibitem[Lin et~al.(2025{\natexlab{a}})Lin, Baltrusaitis, and
  Yekhanin]{lin2025differentiallyworkshop1}
Zinan Lin, Tadas Baltrusaitis, and Sergey Yekhanin.
\newblock Differentially private synthetic data via {API}s 3: Using simulators
  instead of foundation model.
\newblock In \emph{ICLR 2025 Workshop on Navigating and Addressing Data
  Problems for Foundation Models}, 2025{\natexlab{a}}.
\newblock URL \url{https://openreview.net/forum?id=rcPUyG9EOB}.

\bibitem[Lin et~al.(2025{\natexlab{b}})Lin, Baltrusaitis, and
  Yekhanin]{lin2025differentiallyworkshop2}
Zinan Lin, Tadas Baltrusaitis, and Sergey Yekhanin.
\newblock Differentially private synthetic data via {API}s 3: Using simulators
  instead of foundation model.
\newblock In \emph{ICLR 2025 Workshop: Will Synthetic Data Finally Solve the
  Data Access Problem?}, 2025{\natexlab{b}}.
\newblock URL \url{https://openreview.net/forum?id=7AvUO96Pmk}.

\bibitem[Liu et~al.(2024{\natexlab{a}})Liu, Ning, Wang, and
  Lin]{liu2024distilled}
Enshu Liu, Xuefei Ning, Yu~Wang, and Zinan Lin.
\newblock Distilled decoding 1: One-step sampling of image auto-regressive
  models with flow matching.
\newblock \emph{arXiv preprint arXiv:2412.17153}, 2024{\natexlab{a}}.

\bibitem[Liu et~al.(2025)Liu, Chen, Ning, Yan, Dai, Lin, and
  Wang]{liu2025distilled}
Enshu Liu, Qian Chen, Xuefei Ning, Shengen Yan, Guohao Dai, Zinan Lin, and
  Yu~Wang.
\newblock Distilled decoding 2: One-step sampling of image auto-regressive
  models with conditional score distillation.
\newblock \emph{arXiv preprint arXiv:2510.21003}, 2025.

\bibitem[Liu et~al.(2024{\natexlab{b}})Liu, Lyu, Vinaroz, and Park]{dpldm}
Michael~F. Liu, Saiyue Lyu, Margarita Vinaroz, and Mijung Park.
\newblock Differentially private latent diffusion models.
\newblock 2024{\natexlab{b}}.

\bibitem[Liu et~al.(2015)Liu, Luo, Wang, and Tang]{liu2015faceattributes}
Ziwei Liu, Ping Luo, Xiaogang Wang, and Xiaoou Tang.
\newblock Deep learning face attributes in the wild.
\newblock In \emph{Proceedings of International Conference on Computer Vision
  (ICCV)}, December 2015.

\bibitem[McSherry(2009)]{mcsherry2009privacy}
Frank~D McSherry.
\newblock Privacy integrated queries: an extensible platform for
  privacy-preserving data analysis.
\newblock In \emph{Proceedings of the 2009 ACM SIGMOD International Conference
  on Management of data}, pp.\  19--30, 2009.

\bibitem[OpenAI(2023)]{openai2023gpt4}
OpenAI.
\newblock Gpt-4 technical report, 2023.

\bibitem[Riley \& Henderson(2010)Riley and Henderson]{riley2010ns}
George~F Riley and Thomas~R Henderson.
\newblock The ns-3 network simulator.
\newblock In \emph{Modeling and tools for network simulation}, pp.\  15--34.
  Springer, 2010.

\bibitem[Rombach et~al.(2022)Rombach, Blattmann, Lorenz, Esser, and
  Ommer]{rombach2022high}
Robin Rombach, Andreas Blattmann, Dominik Lorenz, Patrick Esser, and Bj{\"o}rn
  Ommer.
\newblock High-resolution image synthesis with latent diffusion models.
\newblock In \emph{Proceedings of the IEEE/CVF Conference on Computer Vision
  and Pattern Recognition}, pp.\  10684--10695, 2022.

\bibitem[Salimans et~al.(2016)Salimans, Goodfellow, Zaremba, Cheung, Radford,
  and Chen]{salimans2016improved}
Tim Salimans, Ian Goodfellow, Wojciech Zaremba, Vicki Cheung, Alec Radford, and
  Xi~Chen.
\newblock Improved techniques for training gans.
\newblock \emph{Advances in neural information processing systems}, 29, 2016.

\bibitem[Schuhmann et~al.(2021)Schuhmann, Vencu, Beaumont, Kaczmarczyk, Mullis,
  Katta, Coombes, Jitsev, and Komatsuzaki]{schuhmann2021laion}
Christoph Schuhmann, Richard Vencu, Romain Beaumont, Robert Kaczmarczyk,
  Clayton Mullis, Aarush Katta, Theo Coombes, Jenia Jitsev, and Aran
  Komatsuzaki.
\newblock Laion-400m: Open dataset of clip-filtered 400 million image-text
  pairs.
\newblock \emph{arXiv preprint arXiv:2111.02114}, 2021.

\bibitem[Singh et~al.(2025)Singh, Fry, Perelman, Tart, Ganesh, El-Kishky,
  McLaughlin, Low, Ostrow, Ananthram, et~al.]{singh2025openai}
Aaditya Singh, Adam Fry, Adam Perelman, Adam Tart, Adi Ganesh, Ahmed El-Kishky,
  Aidan McLaughlin, Aiden Low, AJ~Ostrow, Akhila Ananthram, et~al.
\newblock Openai gpt-5 system card.
\newblock \emph{arXiv preprint arXiv:2601.03267}, 2025.

\bibitem[Sohl-Dickstein et~al.(2015)Sohl-Dickstein, Weiss, Maheswaranathan, and
  Ganguli]{sohl2015deep}
Jascha Sohl-Dickstein, Eric Weiss, Niru Maheswaranathan, and Surya Ganguli.
\newblock Deep unsupervised learning using nonequilibrium thermodynamics.
\newblock In \emph{International Conference on Machine Learning}, pp.\
  2256--2265. PMLR, 2015.

\bibitem[Sun et~al.(2024)Sun, Chen, Gong, Wang, and Li]{sun2024netdpsyn}
Danyu Sun, Joann~Qiongna Chen, Chen Gong, Tianhao Wang, and Zhou Li.
\newblock Netdpsyn: synthesizing network traces under differential privacy.
\newblock In \emph{Proceedings of the 2024 ACM on Internet Measurement
  Conference}, pp.\  545--554, 2024.

\bibitem[Swanberg et~al.(2025)Swanberg, McKenna, Roth, Cheu, and
  Kairouz]{swanberg2025api}
Marika Swanberg, Ryan McKenna, Edo Roth, Albert Cheu, and Peter Kairouz.
\newblock Is api access to llms useful for generating private synthetic tabular
  data?
\newblock \emph{arXiv preprint arXiv:2502.06555}, 2025.

\bibitem[Tao et~al.(2021)Tao, McKenna, Hay, Machanavajjhala, and
  Miklau]{tao2021benchmarking}
Yuchao Tao, Ryan McKenna, Michael Hay, Ashwin Machanavajjhala, and Gerome
  Miklau.
\newblock Benchmarking differentially private synthetic data generation
  algorithms.
\newblock \emph{arXiv preprint arXiv:2112.09238}, 2021.

\bibitem[Towers et~al.(2024)Towers, Kwiatkowski, Terry, Balis, De~Cola, Deleu,
  Goul{\~a}o, Kallinteris, Krimmel, KG, et~al.]{towers2024gymnasium}
Mark Towers, Ariel Kwiatkowski, Jordan Terry, John~U Balis, Gianluca De~Cola,
  Tristan Deleu, Manuel Goul{\~a}o, Andreas Kallinteris, Markus Krimmel, Arjun
  KG, et~al.
\newblock Gymnasium: A standard interface for reinforcement learning
  environments.
\newblock \emph{arXiv preprint arXiv:2407.17032}, 2024.

\bibitem[Tran et~al.(2026)Tran, Backurs, Lin, Reis, Xiong, and
  Yekhanin]{tran2026differentially}
Toan Tran, Arturs Backurs, Zinan Lin, Victor Reis, Li~Xiong, and Sergey
  Yekhanin.
\newblock Differentially private synthetic data via apis 4: Tabular data.
\newblock \emph{arXiv preprint arXiv:2606.08259}, 2026.

\bibitem[Tsai et~al.(2024)Tsai, Li, Chen, Chen, Yu, Ren, and
  Buet-Golfouse]{dplora}
Yu-Lin Tsai, Yizhe Li, Zekai Chen, Po-Yu Chen, Chia-Mu Yu, Xuebin Ren, and
  Francois Buet-Golfouse.
\newblock Differentially private fine-tuning of diffusion models.
\newblock \emph{arXiv preprint arXiv:2406.01355}, 2024.

\bibitem[Vinaroz et~al.(2022)Vinaroz, Charusaie, Harder, Adamczewski, and
  Park]{vinaroz2022hermite}
Margarita Vinaroz, Mohammad-Amin Charusaie, Frederik Harder, Kamil Adamczewski,
  and Mi~Jung Park.
\newblock Hermite polynomial features for private data generation.
\newblock In \emph{International Conference on Machine Learning}, pp.\
  22300--22324. PMLR, 2022.

\bibitem[Wang et~al.(2025{\natexlab{a}})Wang, Lin, Yu, and
  Zhang]{wang2025synthesize}
Haoxiang Wang, Zinan Lin, Da~Yu, and Huishuai Zhang.
\newblock Synthesize privacy-preserving high-resolution images via private
  textual intermediaries.
\newblock \emph{arXiv preprint arXiv:2506.07555}, 2025{\natexlab{a}}.

\bibitem[Wang et~al.(2025{\natexlab{b}})Wang, Raunak, Backurs, Reis, Zhou,
  Chen, Yang, Lin, Yekhanin, and Fanti]{wang2025struct}
Shuaiqi Wang, Vikas Raunak, Arturs Backurs, Victor Reis, Pei Zhou, Sihao Chen,
  Longqi Yang, Zinan Lin, Sergey Yekhanin, and Giulia Fanti.
\newblock Struct-bench: A benchmark for differentially private structured text
  generation.
\newblock \emph{arXiv preprint arXiv:2509.10696}, 2025{\natexlab{b}}.

\bibitem[Wood et~al.(2021)Wood, Baltru{\v{s}}aitis, Hewitt, Dziadzio, Cashman,
  and Shotton]{wood2021fake}
Erroll Wood, Tadas Baltru{\v{s}}aitis, Charlie Hewitt, Sebastian Dziadzio,
  Thomas~J Cashman, and Jamie Shotton.
\newblock Fake it till you make it: face analysis in the wild using synthetic
  data alone.
\newblock In \emph{Proceedings of the IEEE/CVF international conference on
  computer vision}, pp.\  3681--3691, 2021.

\bibitem[Xie et~al.(2024)Xie, Lin, Backurs, Gopi, Yu, Inan, Nori, Jiang, Zhang,
  Lee, et~al.]{xie2024differentially}
Chulin Xie, Zinan Lin, Arturs Backurs, Sivakanth Gopi, Da~Yu, Huseyin~A Inan,
  Harsha Nori, Haotian Jiang, Huishuai Zhang, Yin~Tat Lee, et~al.
\newblock Differentially private synthetic data via foundation model apis 2:
  Text.
\newblock \emph{arXiv preprint arXiv:2403.01749}, 2024.

\bibitem[Xie et~al.(2018)Xie, Lin, and et~al.]{dpgan}
Liyang Xie, Kaixiang Lin, and et~al.
\newblock Differentially private generative adversarial network.
\newblock \emph{CoRR}, abs/1802.06739, 2018.
\newblock URL \url{http://arxiv.org/abs/1802.06739}.

\bibitem[Xie et~al.(2017)Xie, Girshick, Doll{\'a}r, Tu, and
  He]{xie2017aggregated}
Saining Xie, Ross Girshick, Piotr Doll{\'a}r, Zhuowen Tu, and Kaiming He.
\newblock Aggregated residual transformations for deep neural networks.
\newblock In \emph{Proceedings of the IEEE conference on computer vision and
  pattern recognition}, pp.\  1492--1500, 2017.

\bibitem[Xu et~al.(2026)Xu, Chen, Lin, Shi, Graham, Zhou, Wan, Stein,
  Estellers, Chen, et~al.]{xu2026dp}
Fangyuan Xu, Sihao Chen, Zinan Lin, Taiwei Shi, Sydney Graham, Pei Zhou,
  Mengting Wan, Alex Stein, Virginia Estellers, Charles Chen, et~al.
\newblock Dp-rft: Learning to generate synthetic text via differentially
  private reinforcement fine-tuning.
\newblock \emph{arXiv preprint arXiv:2602.18633}, 2026.

\bibitem[Yang et~al.(2023)Yang, Adamczewski, and et~al]{dp-ntk}
Yilin Yang, Kamil Adamczewski, and et~al.
\newblock Differentially private neural tangent kernels for privacy-preserving
  data generation.
\newblock \emph{CoRR}, abs/2303.01687, 2023.

\bibitem[Yin et~al.(2022)Yin, Lin, Jin, Fanti, and Sekar]{yin2022practical}
Yucheng Yin, Zinan Lin, Minhao Jin, Giulia Fanti, and Vyas Sekar.
\newblock Practical gan-based synthetic ip header trace generation using
  netshare.
\newblock In \emph{Proceedings of the ACM SIGCOMM 2022 Conference}, pp.\
  458--472, 2022.

\bibitem[Yu et~al.(2021)Yu, Naik, Backurs, Gopi, Inan, Kamath, Kulkarni, Lee,
  Manoel, Wutschitz, et~al.]{yu2021differentially}
Da~Yu, Saurabh Naik, Arturs Backurs, Sivakanth Gopi, Huseyin~A Inan, Gautam
  Kamath, Janardhan Kulkarni, Yin~Tat Lee, Andre Manoel, Lukas Wutschitz,
  et~al.
\newblock Differentially private fine-tuning of language models.
\newblock \emph{arXiv preprint arXiv:2110.06500}, 2021.

\bibitem[Yu et~al.(2023)Yu, Gopi, Kulkarni, Lin, Naik, Religa, Yin, and
  Zhang]{yu2023selective}
Da~Yu, Sivakanth Gopi, Janardhan Kulkarni, Zinan Lin, Saurabh Naik,
  Tomasz~Lukasz Religa, Jian Yin, and Huishuai Zhang.
\newblock Selective pre-training for private fine-tuning.
\newblock \emph{arXiv preprint arXiv:2305.13865}, 2023.

\bibitem[Yue et~al.(2022)Yue, Inan, Li, Kumar, McAnallen, Sun, Levitan, and
  Sim]{yue2022synthetic}
Xiang Yue, Huseyin~A Inan, Xuechen Li, Girish Kumar, Julia McAnallen, Huan Sun,
  David Levitan, and Robert Sim.
\newblock Synthetic text generation with differential privacy: A simple and
  practical recipe.
\newblock \emph{arXiv preprint arXiv:2210.14348}, 2022.

\bibitem[Zagoruyko \& Komodakis(2016)Zagoruyko and
  Komodakis]{zagoruyko2016wide}
Sergey Zagoruyko and Nikos Komodakis.
\newblock Wide residual networks.
\newblock \emph{arXiv preprint arXiv:1605.07146}, 2016.

\bibitem[Zou et~al.(2025)Zou, Liu, Li, Xiong, Zhang, Liu, Ye, Ouyang, and
  Zhang]{zou2025contrastive}
Tianyuan Zou, Yang Liu, Peng Li, Yufei Xiong, Jianqing Zhang, Jingjing Liu,
  Xiaozhou Ye, Ye~Ouyang, and Ya-Qin Zhang.
\newblock Contrastive private data synthesis via weighted multi-plm fusion.
\newblock \emph{arXiv preprint arXiv:2502.00245}, 2025.

\bibitem[Zou et~al.(2026)Zou, Lin, Gopi, Liu, Zhang, Sim, Deng, and
  Yekhanin]{zou2026pesgd}
Tianyuan Zou, Zinan Lin, Sivakanth Gopi, Yang Liu, Ya-Qin Zhang, Robert Sim,
  Xin Deng, and Sergey Yekhanin.
\newblock {PE}-{SGD}: Differentially private deep learning via evolution of
  gradient subspace for text.
\newblock In \emph{The Fourteenth International Conference on Learning
  Representations}, 2026.
\newblock URL \url{https://openreview.net/forum?id=713ywmTZHv}.

\end{thebibliography}
\bibliographystyle{tmlr}

\clearpage

\appendix

\part{Appendix}
\etocsetnexttocdepth{subsection}
\localtableofcontents

\clearpage

\FloatBarrier
\section{Private Evolution}
\label{app:pe}

\cref{alg:main_full} presents the \privateevolution{} (\pe{}) algorithm, reproduced from \citet{lin2023differentially}.  
This algorithm represents the conditional version of \pe{}, where each generated image is associated with a class label.  
It can be interpreted as running the unconditional version of \pe{} separately for each class (\cref{line:for_loop}).  

\begin{algorithm}[thpb]
    \DontPrintSemicolon
    \LinesNumbered
	\BlankLine
	\SetKwInOut{Input}{Input}
	\caption{\privateevolution{} (\pe{})}
    \label{alg:main_full}
	\Input{The set of private classes: $\privatesampleclassset$ ~~~ 
 ($\privatesampleclassset=\brc{0}$ if for unconditional generation)\\
 Private samples: $\privatesampleset=\brc{(x_i, y_i)}_{i=1}^{\numprisamples}$, where $x_i$ is a sample and $y_i\in \privatesampleclassset$ is its label\\
 Number of iterations: $\numiterations$\\
 Number of generated samples: $\numgensamples$~~~(assuming $\numgensamples~\textrm{mod}~\brd{\privatesampleclassset}=0$)\\
 Noise multiplier for \dpvotingname{}: $\noisemultiplier$\\
 Threshold for \dpvotingname{}: $\threshold$
	}
	\BlankLine
        $\generatedsampleset \leftarrow \emptyset$\;
        \For{$c \in \privatesampleclassset$}{ \label{line:for_loop}
            $private\_samples \leftarrow \brc{x_i| (x_i, y_i)\in \privatesampleset \text{  and  } y_i=c}$ \label{line:loop_start}\;
            $S_0 \leftarrow \randomsampleapi{\numgensamples / \brd{\privatesampleclassset}}$ \; \label{line:initial}
    	\For{$t \leftarrow 1, \ldots, \numiterations$}
    	{
                $histogram_t \leftarrow \dpvotingfunction{private\_samples,S_{t-1}, \noisemultiplier,\threshold}$ \label{line:voting} \tcp*{See \cref{alg:voting}} 
                $\calP_t \leftarrow histogram_t/\mathrm{sum}(histogram_t)$ \label{line:normalization} \tcp*{$\calP_t$ is a distribution on $S_t$}
                $S_{t}'\leftarrow $ draw $N_\syn/|C|$ samples with replacement from $\calP_t$ \label{line:draw} \tcp*{$S_t'$ is a multiset}
                $S_{t}\leftarrow\samplevariationapi{S_t'}$  \label{line:variation}
            }
            $\generatedsampleset\leftarrow \generatedsampleset \cup \brc{(x,c)|x\in S_{T}}$ \label{line:loop_end}
        }
	\Return{ $\generatedsampleset$}
\end{algorithm}
\begin{algorithm}[thpb]
    \DontPrintSemicolon
    \LinesNumbered
	\BlankLine
	\SetKwInOut{Input}{Input}
	\SetKwInOut{Output}{Output}
	\caption{\dpvotingname{} (\dpvotingfunctionname{})}
    \label{alg:voting}
	\Input{Private samples: $S_{\priv}$\\
 Generated samples: $S=\brc{z_i}_{i=1}^{n}$\\
 Noise multiplier: $\noisemultiplier$\\
 Threshold: $\threshold$\\
 Distance function: $\distancefunction{\cdot,\cdot}$
	}
 \Output{DP nearest neighbors histogram on $S$}
	\BlankLine
        $histogram\leftarrow[0,\ldots,0]$\label{line:hist_init}\; 
        \For{$x_{\priv} \in S_{\priv}$\label{line:hist_for}}{ 
            $i=\arg\min_{j\in\brb{n}} \distancefunction{x_{\priv},z_j}$ \label{line:distance}\;
            $histogram[i] \leftarrow histogram[i] + 1$ \label{line:hist_assign}
        }
        $histogram \leftarrow histogram + \normaldistribution{0}{\noisemultiplier I_n}$ \label{line:add_noise}\tcp*{Add noise to ensure DP}
        $histogram \leftarrow \max\bra{histogram -\threshold, 0}$ \tcp*{`max', `-' are element-wise} \label{line:threshold}
	\Return{ $histogram$}
\end{algorithm}

\FloatBarrier
\section{Theoretical Analysis}
\label{app:theoretical}

\myparatightestn{Convergence analysis.}
Since \simpe{} only changes \pe{}’s \randomsampleapiname{} and \samplevariationapiname{}, the original convergence analysis (App. E in \citet{lin2023differentially}) remains valid for \simpe{}. Specifically:
\begin{packeditemize}
\item The analysis assumes that \randomsampleapiname{} generates samples within a ball of diameter $D$ covering the private samples. Changing \randomsampleapiname{} affects $D$ but not the overall analysis procedure.
\item  The analysis assumes that \samplevariationapiname{} draws samples from a Gaussian distribution—an essential assumption for any tractable analysis involving complex, unknown foundation models and simulators. Under this assumption, modifying \samplevariationapiname{} does not impact the analysis procedure.
\end{packeditemize}

\myparatightestn{Privacy analysis.} Since \simpe{} does not change the way \pe{} utilizes private data, the privacy analysis of \pe{} \cite{lin2023differentially} applies to \simpe{}. We repeat the privacy analysis in \pe{} below for completeness.

The only step that directly accesses private data is \dpvotingfunctionname{} (\cref{line:voting} in \cref{alg:main_full}, detailed in \cref{alg:voting}). Let us analyze its privacy cost first:

\begin{packeditemize}
    \item Each private sample contributes only one vote (\cref{line:add_noise} in \cref{alg:voting}).
    \item Adding or removing a single sample alters the resulting histogram by at most 1 in the $l_2$ norm—so the sensitivity is 1.
    \item \cref{line:add_noise} adds i.i.d. Gaussian noise with standard deviation $\noisemultiplier$ to each histogram bin. This corresponds to the standard Gaussian mechanism \cite{dwork2014algorithmic} with noise multiplier $\noisemultiplier$ (i.e., $\noisemultiplier/1$).
\end{packeditemize}

Now, consider the entire \simpe{} procedure (\cref{alg:main_full}). 

\emph{(1) Let's first consider the procedure for one class $c$ (\cref{line:loop_start}-\cref{line:loop_end}).} All steps other than \dpvotingfunctionname{} do not access private data and simply post-process results from earlier DP steps. Therefore, \cref{alg:main_full} can be interpreted as \emph{$\numiterations$ adaptive compositions} of Gaussian mechanisms.

According to Corollary 3.3 in \citet{dong2022gaussian},\footnote{Please see their arXiv version \url{https://arxiv.org/pdf/1905.02383}.} \emph{$\numiterations$ compositions of Gaussian mechanisms} with noise multiplier $\noisemultiplier$ can be viewed as a \emph{single Gaussian mechanism} with effective noise multiplier $\noisemultiplier/\sqrt{\numiterations}$.

Thus, the final privacy cost reduces to computing the $\varepsilon$ and $\delta$ of a standard Gaussian mechanism with noise $\noisemultiplier/\sqrt{\numiterations}$. We use the tight results from \citet{balle2018improving} for this computation: the mechanism is $(\varepsilon, \delta)$-differentially private if
\[
\Phi\!\left( \frac{\sqrt{T}}{2\sigma} - \frac{\varepsilon \sigma}{\sqrt{T}} \right)
- e^{\varepsilon} \, \Phi\!\left( -\frac{\sqrt{T}}{2\sigma} - \frac{\varepsilon \sigma}{\sqrt{T}} \right)
\leq \delta
\]
where $\Phi$ denotes the Gaussian cumulative distribution function (CDF). Given $\delta$ and $\noisemultiplier$, we can use the above equation to solve for $\varepsilon$.

\emph{(2) Now, let's first consider the full algorithm involving multiple classes.} Since the private samples from different classes are disjoint, due to \emph{parallel composition property} \cite{mcsherry2009privacy}, the privacy cost remains the same as above. See \citet{lin2023differentially} for an alternative analysis that gives the same conclusion.

\FloatBarrier
\section{Justifying the Methodological Design of \simpe{} with Simulator-generated Data}
\label{app:justification_data_pe}

In this section, we provide experimental evidence to support the claim in \cref{sec:method_data}: ``If we already know that a sample $\simdata_i$ is far from the private dataset, then its nearest neighbors in $\simdataset$ are also likely to be far from the private dataset.''

Consider \simpe{}'s synthetic dataset $S_t$ from the $t$-th iteration. For each private sample $x_i$ in the private dataset $\privatesampleset{}$, we find its nearest sample in $S_t$ as $z_i$. 
$S(selected)_t=\brc{z_i}_{i=1}^{\numprisamples{}}$ 
contains samples close to $\privatesampleset{}$ (chosen by \dpvotingfunctionname{} (\cref{alg:voting}) in PE under non-DP settings). Conversely, the remaining samples, $S(unselected)_t=S_t\setminus S(selected)_t$ 
are farther from $\privatesampleset{}$. 

Now, we need to confirm that  $S(unselected)_t$'s nearest neighbors in $\simdataset{}$ are farther away from the private dataset $\privatesampleset{}$ than $S(selected)_t$. We define:
\begin{packeditemize}
  \item $z(Y)_i$ as the nearest neighbor of the private sample $x_i$ in $S(Y)_t$, for $Y\in\brc{selected, unselected}$
  \item $q(Y)_j^i$ as the $j-th$ nearest neighbor of $z(Y)_i  \text{ in } \simdataset$
  \item $FID(Y)$ as the FID between $S_{priv}$ and $\brc{q(Y)_j^i}_{i=1,j=1}^{\numprisamples,\nnvariationdegree}$. 
\end{packeditemize}

\begin{figure}[h]
    \centering
    \includegraphics[width=0.3\linewidth]{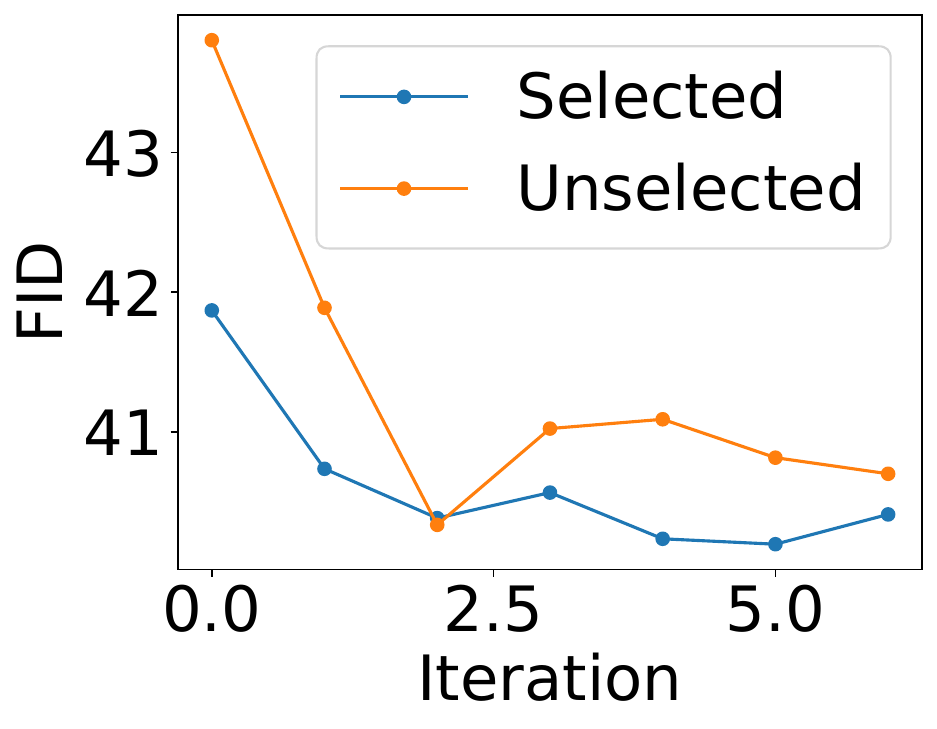}
    \caption{FID of the nearest neighbors of the selected and unselected samples.}
    \label{fig:ablation_verify_non_voted_samples_farther_from_private_data}
\end{figure}

\cref{fig:ablation_verify_non_voted_samples_farther_from_private_data} shows that $FID(selected)$ is smaller (better) than $FID(unselected)$ across most PE iterations, supporting our claim.

\FloatBarrier
\section{Experimental Details}
\label{app:exp_details}

In this section, we provide more experimental details.

\subsection{\mnist{} with Text Rendering Program}
\label{app:mnist_exp_details}

\textbf{The categorical parameters include:}
\underline{(1) Font.} We use Google Fonts \cite{googlefonts}, which offers 3589 fonts in total.
\underline{(2) Text.} The text consists of digits `0' - `9'. %
\textbf{The numerical parameters include:}
\underline{(1) Font size}, ranging from 10 to 29.
\underline{(2) Stroke width}, ranging from 0 to 2.
\underline{(3) Digit rotation degree}, ranging from $-30\degree$ to $30\degree$. 

\cref{tab:hyper_mnist_cate,tab:hyper_mnist_nume} show %
their %
variation degrees. %
The total number of \pe{} iterations is 4. Following \citet{dpimagebench}, we set the number of generated samples to be 60,000.

\begin{table}[!h]
    \centering
    \caption{The configurations of the categorical parameters in \mnist{} with Text Rendering Program experiments.}
    \label{tab:hyper_mnist_cate}
    \begin{tabular}{c|c|c}
    \toprule
       Categorical Parameter ($\cate$)  & Feasible Set ($\cateset$)  & Variation Degrees ($\catevariationdegree$) across \pe{} Iterations\\\midrule
       Font  & 1 - 3589 & 0.8, 0.4, 0.2, 0.0\\
       Text & `0' - `9' & 0, 0, 0, 0\\
         \bottomrule
    \end{tabular}
\end{table}

\begin{table}[!h]
    \centering
    \caption{The configurations of the numerical parameters in \mnist{} with Text Rendering Program experiments.}
    \label{tab:hyper_mnist_nume}
    \begin{tabular}{c|c|c}
    \toprule
       Numerical Parameter ($\nume$)  & Feasible Set ($\numeset$)  & Variation Degrees ($\numevariationdegree$) across \pe{} Iterations\\\midrule
       Font size  & [10, 30] & 5, 4, 3, 2\\
       Font rotation & [-30, 30] & 9, 7, 5, 3\\
       Stroke width & [0, 2] & 1, 1, 0, 0\\
         \bottomrule
    \end{tabular}
\end{table}

\subsection{\celeba{} with Generated Images from Computer Graphics-based Render}
\label{app:avatar_exp_details}

The variation degrees $\nnvariationdegree$ across \pe{} iterations are [1000, 500, 200, 100, 50, 20]. The total number of \pe{} iterations is 6. Following \citet{dpimagebench}, we set the number of generated samples to be 60,000.

\subsection{\celeba{} with Rule-based Avatar Generator}

The full list of the categorical parameters are 
    \begin{packeditemize}
    \item Style
    \item Background color
    \item Top
    \item Hat color
    \item Eyebrows
    \item Eyes
    \item Nose
    \item Mouth
    \item Facial hair
    \item Skin color
    \item Hair color
    \item Facial hair color
    \item Accessory
    \item Clothing
    \item Clothing color
    \item Shirt graphic
\end{packeditemize}
These are taken from the input parameters to the library \cite{pythonavatar}. There is no numerical parameter. 

For the experiments with only the simulator,
the variation degrees $\catevariationdegree$ across \pe{} iterations are [0.8, 0.6, 0.4, 0.2, 0.1, 0.08, 0.06].
The total number of \pe{} iterations is 7. Following \citet{dpimagebench}, we set the number of generated samples to be 60,000.

For the experiments with both foundation models and the simulator, we use a total of 5 \pe{} iterations so as to be consistent with the setting in \citet{dpimagebench}.
For the \randomsampleapiname{} and the first \pe{} iteration, we use the simulator ($\catevariationdegree=0.8$). For the next 4 \pe{} iterations, we use the same foundation model as in \citet{lin2023differentially} with variation degrees [96, 94, 92, 90]. Following \citet{dpimagebench}, we set the number of generated samples to be 60,000.

\subsection{The CartPole Environment}
\label{sec:app_cartpole_settings}

\myparatightestn{Policy.}
To map observations to control signals, we adopt a simple linear policy of the form
\[
\mathrm{sign}(\langle \text{observations}, \boldsymbol{\pi} \rangle),
\]
where $\boldsymbol{\pi}$ is a 4-dimensional policy vector and $\langle \cdot, \cdot \rangle$ denotes the inner product.

\myparatightestn{Private Dataset.}
To create a ``private dataset,'' we enumerate all policies with each coordinate in
$\{-5, -4.5, \ldots, 5\}$, resulting in a total of $21^4$ policies. We then uniformly
sample 2500 policies that succeed (label $=1$) and 2500 policies that fail (label $=0$).

\myparatightestn{Embedding.}
We use raw observations as the embedding. Since episodes may terminate at different
lengths (due to success or failure), we pad all observation sequences to a fixed length
of 500 frames by repeating the final observation. To keep the setup simple, we use the
raw values directly without any normalization, and we find that this simple embedding
already yields strong results. This also highlights that \simpe{} can still function effectively without
any pretrained embeddings.

\myparatightestn{RANDOM and VARIATION APIs.}
We follow the same method as in \cref{sec:method_simulator}:
\begin{itemize}
    \item \randomsampleapiname{} samples policy parameters uniformly at random.
    \item \samplevariationapiname{} perturbs existing policy parameters.
\end{itemize}

\FloatBarrier
\section{Additional Results}

\subsection{\simpe{} with Weak Simulators}
\label{app:weak_simulator}

\begin{figure*}[!t]
\centering
\vspace{-0cm}
\begin{minipage}{0.32\textwidth}
  \centering
\includegraphics[width=0.95\textwidth]{fig/6_celeba_images/image_sample/000000000.png}
\subcaption{Real (private) images}\label{fig:celeba_avatar_real}
\end{minipage}%
\begin{minipage}{0.32\textwidth}
  \centering
\includegraphics[width=0.95\textwidth]{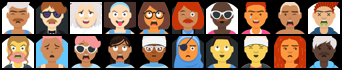}
\subcaption{Simulator-generated images}\label{fig:celeba_avatar_simulator}
\end{minipage}%
\begin{minipage}{0.32\textwidth}
  \centering
\includegraphics[width=0.95\textwidth]{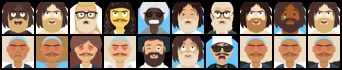}
\subcaption{\simpe{} images ($\epsilon=10$)}\label{fig:celeba_avatar_simpe}
\end{minipage}
\vspace{-0.2cm}
\caption{The real and generated images on \celeba{}. The simulator is a weak rule-based avatar generator \cite{pythonavatar} significantly different from the real dataset. The top rows correspond to the ``female'' class, and the bottom rows correspond to the ``male'' class. 
The simulator generates images with incorrect classes. \simpe{} tends to generate faces with long hair for the female class and short hair for the male class (correctly), but the generated images have mode collapse issues.} \label{fig:celeba_avatar}
\vspace{-0.4cm}
\end{figure*}

In \cref{fig:celeba_avatar_simpe}, we see that \simpe{}'s generated images have many duplicates. In this section, we conduct more analysis on this phenomenon.

\begin{figure}[h]
    \centering
    \includegraphics[width=0.3\linewidth]{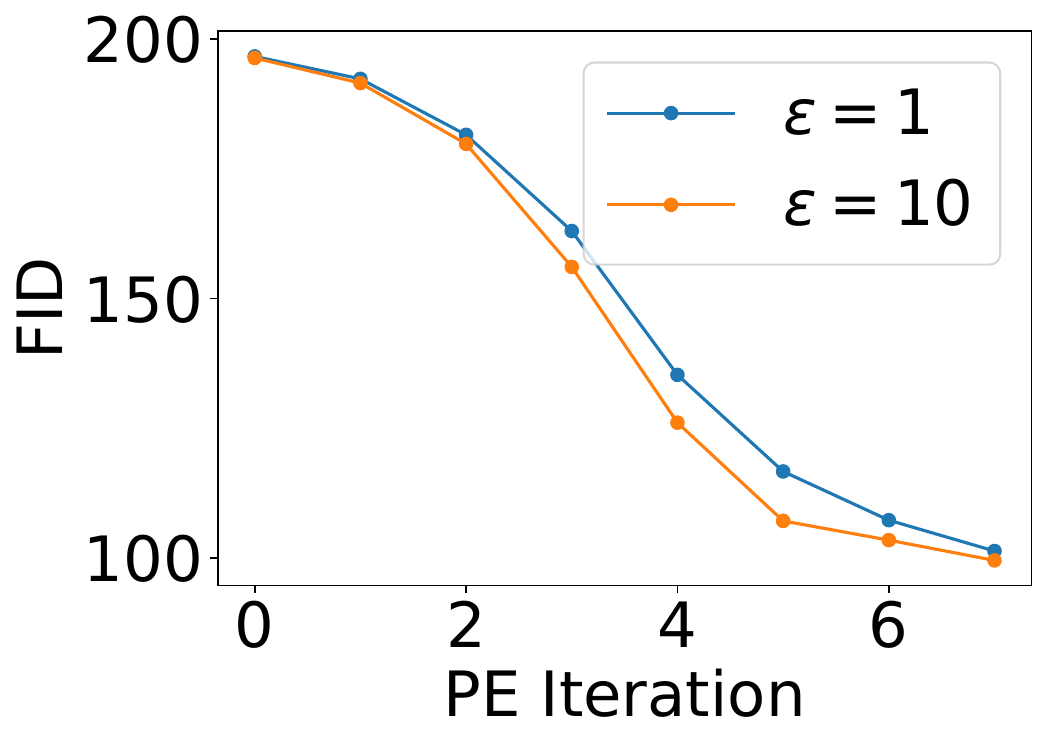}
    \caption{\simpe{} (using \digiface{})'s FID on \celeba{} improves over the course of the \pe{} iterations.}
    \label{fig:fid_datape_celeba}
\end{figure}

\begin{figure}[h]
    \centering
    \includegraphics[width=0.3\linewidth]{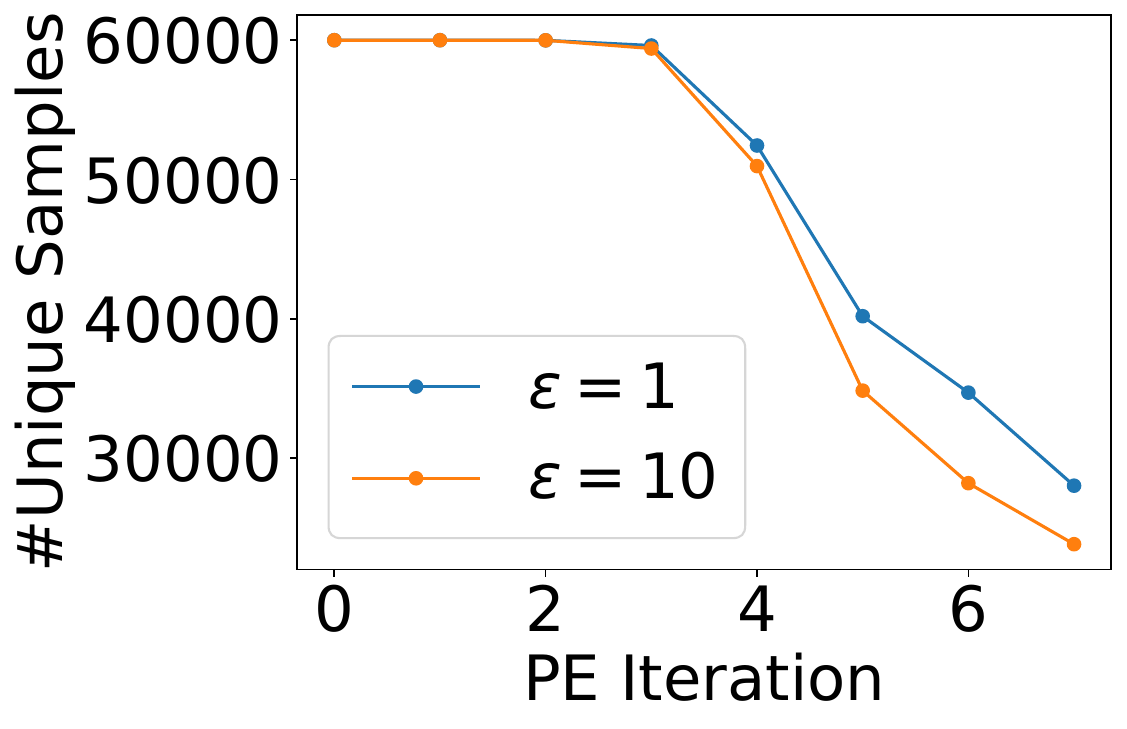}
    \caption{The number of unique generated samples in each iteration of \simpe{} (using \digiface{}) on \celeba{}. }
    \label{fig:unique_datape_celeba}
\end{figure}

\cref{fig:fid_datape_celeba} shows that FID between generated images and the private dataset decreases (improves) across iterations, confirming that generated images better align with the private distribution over iterations. However, the number of unique samples in the generated dataset drops significantly (\cref{fig:unique_datape_celeba}), leading to duplicates in \cref{fig:celeba_avatar_simpe}. This may be due to the simulator being too weak, generating most images far from the private dataset. As a result, \simpe{} converges to a small set of good images.

\FloatBarrier
\subsection{\simpe{} with \cifar{} as the Private Dataset and \imagenet{} as the Simulator-generated Dataset}
\label{app:cifar10}

In this section, we apply the algorithm in \cref{sec:method_data} on the setting where \cifar{} is treated as the private dataset and \imagenet{} is treated as the simulator-generated dataset. We conduct this experiment for the following reasons:
\begin{packeditemize}
    \item In \cref{sec:method_data}, we discuss that \simpe{}  can apply to any public dataset beyond simulator-generated data. This experiment provides evidence for that claim.
    \item The original PE approach on \cifar{} \cite{lin2023differentially} uses a foundation model pretrained on ImageNet, which aligns with the simulator data in this experiment. Comparing their performance removes the impact of distribution alignment and highlights the differences in the \randomsampleapiname{} and \samplevariationapiname{} designs of \pe{} and \simpe{}.
\end{packeditemize}

\myparatightestn{Experiment settings.} 
We use $\epsilon=10$. The downstream task is classifying images into the 10 classes in CIFAR10, following DPImageBench \cite{dpimagebench} and the original PE paper \cite{lin2023differentially}.

\myparatightestn{Results.}  \simpe{} achieves FID=8.76, and acc=70.06\%. In comparison, \pe{} with a foundation model achieves similarly with FID=9.2 and acc=75.3\%. These results suggest that \simpe{} and \pe{} can be competitive when both suitable simulators and foundation models are available for the private dataset. In addition, the results are comparable to state-of-the-art training-based methods, demonstrating that \simpe{} is a promising framework.

\FloatBarrier
\subsection{\revisiontmlr{\simpe{} for Generating Networking Data }}
\label{app:network}

\revisiontmlr{In our main paper, we demonstrated that \simpe{} works for images and robotics simulations. In this section, we additionally provide a prelimiality expeirments to show that \simpe{} can also be used for generating networking data.}

\revisiontmlr{We follow the experimental setup of \citet{sun2024netdpsyn}, a recent work on DP networking data synthesis.}

\revisiontmlr{\myparatightestn{Data.}
We aim to generate differentially private \emph{network flow header} data using the TON dataset~\citep{sun2024netdpsyn} as the private dataset. Each sample contains the following fields:
\begin{packeditemize}
    \item Source IP address
    \item Destination IP address
    \item Source port
    \item Destination port
    \item Protocol (e.g., UDP, TCP)
    \item Timestamp of the first packet
    \item Duration of the flow
    \item Number of packets
    \item Number of bytes
    \item Flow label (10 classes, e.g., normal, background, DDoS)
\end{packeditemize}}

\revisiontmlr{\myparatightestn{RANDOM and VARIATION APIs.}
We follow the same method as in \cref{sec:method_simulator}:
\begin{itemize}
    \item \randomsampleapiname{} generates random flows.
    \item \samplevariationapiname{} perturbs individual fields of a flow (e.g., modifying the IP address).
\end{itemize}}

\revisiontmlr{\myparatightestn{Embedding.}
To keep the setup minimal, we embed samples directly in their normalized raw format, using one-hot encodings for categorical fields and normalized values for continuous fields. Thus, \textbf{the embedding does not rely on any machine learning model}.}

\revisiontmlr{\myparatightestn{Metrics.}
Following~\citep{sun2024netdpsyn}, we report:
\begin{packeditemize}
    \item \textbf{1-way and 2-way Total Variation Distance (TVD)} between synthetic and private data (lower is better).
    \item \textbf{Classification accuracy}, obtained by training a classifier on synthetic data and testing on held-out private data (higher is better).
\end{packeditemize}
This classification task represents a standard downstream application for network flow data.}

\revisiontmlr{\myparatightestn{Privacy budget.}
As in~\citep{sun2024netdpsyn}, we set $\epsilon = 2$.}

\revisiontmlr{\myparatightestn{Results.}
\Cref{tab:network_results} shows how the metrics evolve across \simpe{} iterations. We highlight two key observations:
\begin{packeditemize}
    \item \textbf{Consistent improvement across iterations.} All metrics steadily improve, with final performance significantly better than the initial random samples (e.g., accuracy improves by \textbf{16.76\%}). This demonstrates that \simpe{} is effective for networking data even without ML-based embeddings or sophisticated simulators.
    \item \textbf{Strong performance compared to prior work.} Compared to Figure~3 in~\citet{sun2024netdpsyn}, this simple approach already outperforms two baselines in accuracy. Notably, NetShare \cite{yin2022practical}, a widely used DP fine-tuning--based synthesizer in networking, is surpassed by \simpe{} in classification accuracy, highlighting its potential beyond the image domain.
\end{packeditemize}
}

\begin{table}[t]
\centering
\small
\begin{tabular}{c|c|c|c}
\toprule
Iteration & Accuracy & 1-way TVD & 2-way TVD \\
\midrule
0  & 56.48 & 0.29 & 0.46 \\
1  & 56.48 & 0.29 & 0.45 \\
2  & 56.48 & 0.28 & 0.44 \\
3  & 56.75 & 0.27 & 0.43 \\
4  & 57.08 & 0.27 & 0.42 \\
5  & 58.28 & 0.26 & 0.41 \\
6  & 60.29 & 0.25 & 0.40 \\
7  & 61.28 & 0.24 & 0.38 \\
8  & 63.47 & 0.23 & 0.37 \\
9  & 65.95 & 0.22 & 0.36 \\
10 & 67.37 & 0.21 & 0.35 \\
11 & 68.47 & 0.21 & 0.33 \\
12 & 72.18 & 0.20 & 0.32 \\
13 & 72.77 & 0.19 & 0.31 \\
14 & 73.24 & 0.18 & 0.30 \\
\bottomrule
\end{tabular}
\caption{\revisiontmlr{Performance of \simpe{} on networking data across iterations.}}
\label{tab:network_results}
\end{table}
\FloatBarrier
\section{Additional Ablation Studies}
\label{app:ablation}

\subsection{Data Selection Algorithms}
\label{app:exp_data_selection}

In \cref{sec:method_data}, we discussed two simple alternatives for simulator data selection. The comparison is in \cref{tab:celeba_data_selection_baseline}. We see that \simpe{} with iterative data selection outperforms the baselines on most metrics, validating the intuition outlined in \cref{sec:method_data}. However, the clustering approach used in the second baseline still has merit, as it results in a better FID for $\epsilon=10$. This idea is orthogonal to the design of \simpe{} and could potentially be combined for further improvement. We leave this exploration to future work.

\begin{table}[ht]

    \centering
    \caption{
    Accuracy (\%) of classifiers trained on synthetic images and FID of synthetic images on \celeba{}. The best results are highlighted in bold. \simpe{} outperforms the baselines in most metrics.}
    \label{tab:celeba_data_selection_baseline}
    \vspace{-0.0cm}
    \resizebox{0.6\linewidth}{!}{
    \begin{tabular}{l|cc|cc}
    \toprule
    \multirow{2}{*}{Algorithm} & \multicolumn{2}{c|}{{FID $\downarrow$}} & \multicolumn{2}{c}{{Classification Acc. $\uparrow$}}\\
    \cline{2-5}
     & $\epsilon = 1$ & $\epsilon = 10$ &  $\epsilon = 1$ & $\epsilon = 10$  \\
     \hline
    \dpvotingfunctionname{} on $\generatedsampleset$ & 36.2 & 29.3 & 61.5 & 71.9 \\
    \dpvotingfunctionname{} on cluster centers of $\generatedsampleset$ & 26.4 & \textbf{18.3} & 74.7 & 77.7 \\
    \simpe{} & \textbf{24.7} & 20.8 & \textbf{80.0} & \textbf{82.5} \\
    \bottomrule
\end{tabular}}
\vspace{-0.cm}
\end{table}

\FloatBarrier

\subsection{Performance across Iterations}

\begin{figure*}[!t]
\centering
\begin{minipage}{0.34\textwidth}
  \centering
\includegraphics[width=0.95\textwidth]{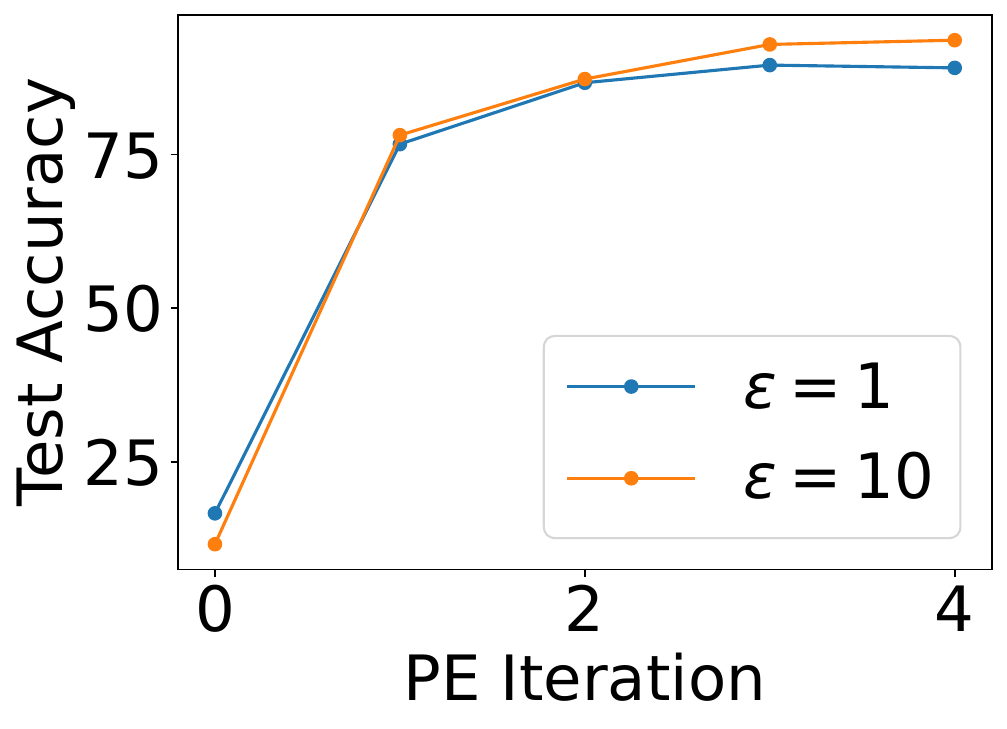}
\subcaption{Acc. on \mnist{}}
\end{minipage}%
\begin{minipage}{0.34\textwidth}
  \centering
\includegraphics[width=0.95\textwidth]{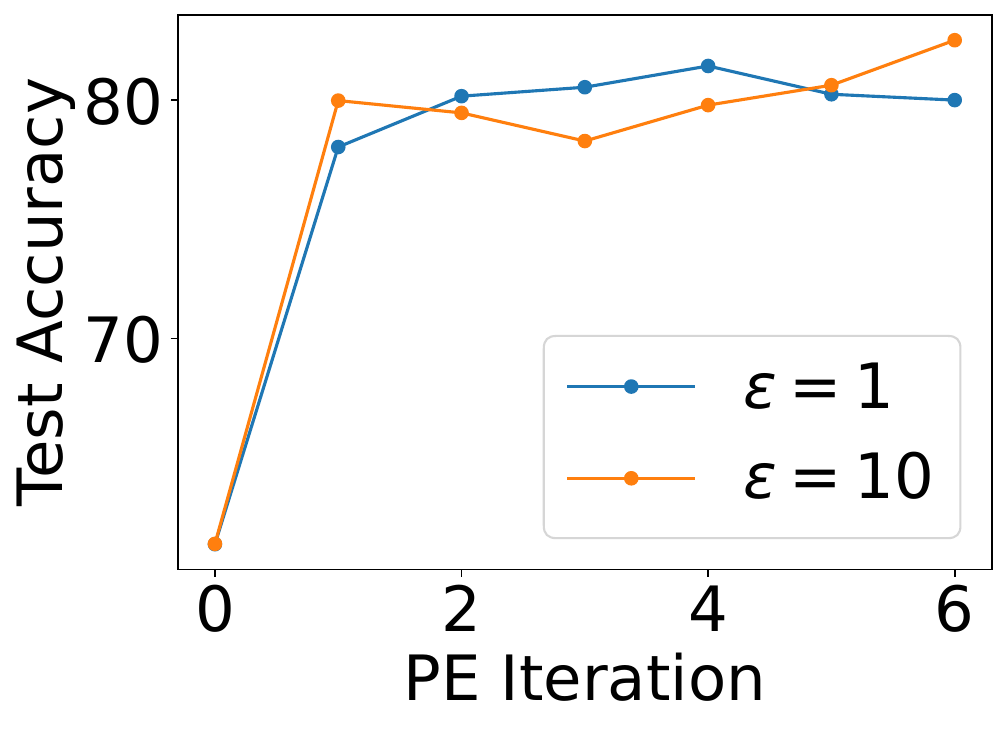}
\subcaption{Acc. on \celeba{}}
\end{minipage}
\begin{minipage}{0.34\textwidth}
  \centering
\includegraphics[width=0.95\textwidth]{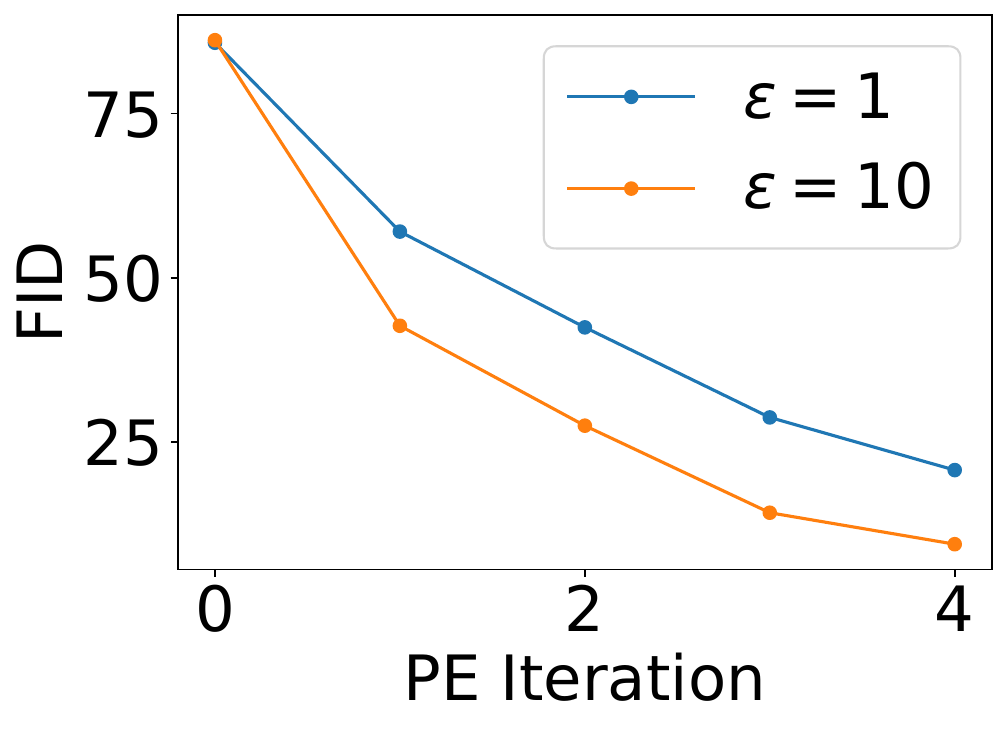}
\subcaption{FID on \mnist{}}
\end{minipage}
\begin{minipage}{0.34\textwidth}
  \centering
\includegraphics[width=0.95\textwidth]{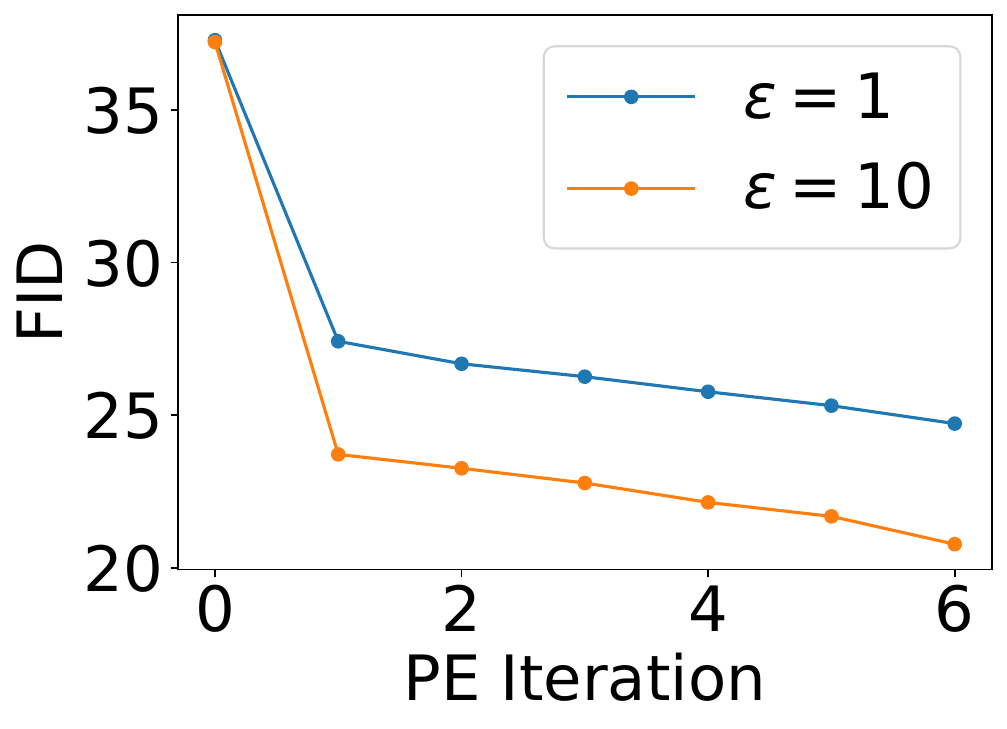}
\subcaption{FID on \celeba{}}
\end{minipage}
\caption{\simpe{}'s FID and accuracy generally improve over the course of the iterations.} \label{fig:metrics_vs_iterations}
\vspace{-0.2cm}
\end{figure*}

\cref{fig:metrics_vs_iterations} shows that both the FID and the downstream classifier's accuracy generally improve as \pe{} progresses. This confirms that \pe{}'s iterative data refinement process is effective when combined with simulators.

\FloatBarrier

\subsection{Sensitivity Analysis of $\numevariationdegree, \catevariationdegree, \nnvariationdegree$}

These hyperparameters (defined in \cref{sec:method_simulator,sec:method_data}) control the variation degree in \samplevariationapiname{}. Higher values induce larger variations. \textbf{These are the only new hyperparameters introduced in \simpe{}.}

\begin{figure*}[h]
\centering
\begin{minipage}{0.49\textwidth}
  \centering
\includegraphics[width=0.7\textwidth]{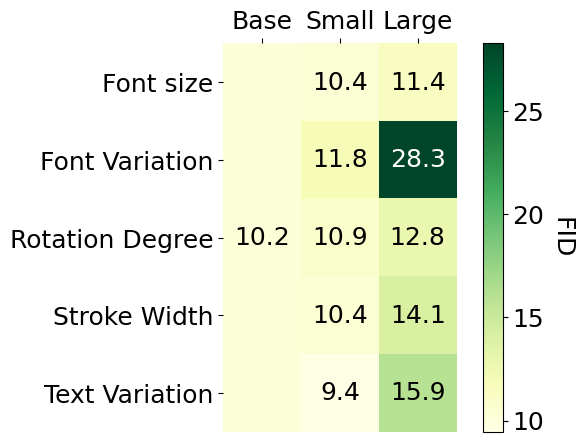}
\subcaption{FID}
\end{minipage}%
\begin{minipage}{0.49\textwidth}
  \centering
\includegraphics[width=0.7\textwidth]{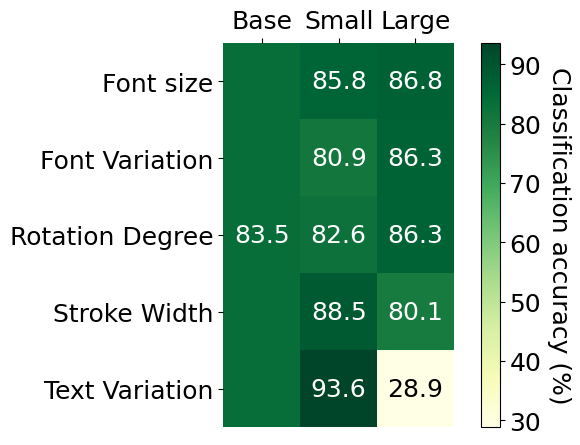}
\subcaption{Classification accuracy}
\end{minipage}%
\caption{Accuracy and FID on \mnist{}
with different $\numevariationdegree,\catevariationdegree$ schedules ($\epsilon=10$). The ``base'' parameter schedule across iterations is: (a) font size's $\numevariationdegree=[5, 4, 3, 2]$, (b) rotation degree's $\numevariationdegree=[9, 7, 5, 3]$, (c) stroke width's $\numevariationdegree=[1, 1, 0, 0]$, (d) font's $\catevariationdegree=[0.8, 0.4, 0.2, 0.0]$, (e) text's $\catevariationdegree=[0.8, 0.4, 0.2, 0.0]$. ``Small'' (``large'') means fixing $\numevariationdegree/\catevariationdegree$ to be the smallest (largest) value across iterations. For example, the unit under ``font variation'' and ``large'' indicates the results when we set font's $\catevariationdegree=[0.8,0.8,0.8,0.8]$.  } \label{fig:ablation_alpha_beta}
\end{figure*}

\begin{figure*}[h]
\centering
\begin{minipage}{0.49\textwidth}
  \centering
\includegraphics[width=0.7\textwidth]{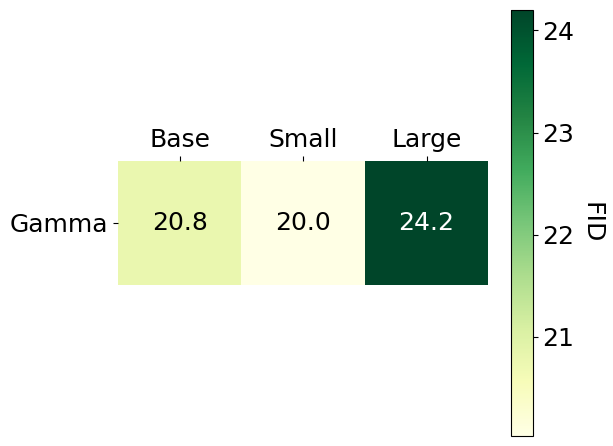}
\subcaption{FID}
\end{minipage}%
\begin{minipage}{0.49\textwidth}
  \centering
\includegraphics[width=0.7\textwidth]{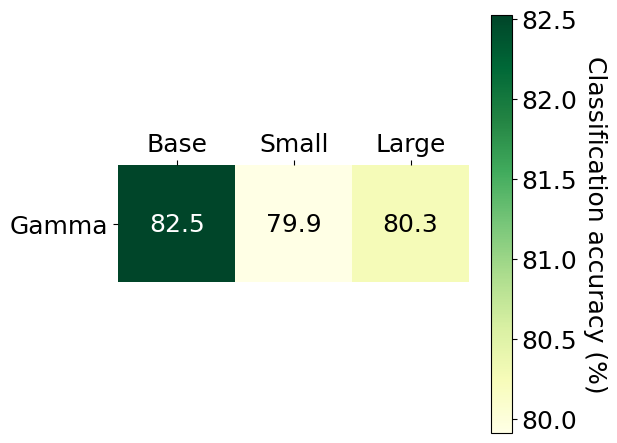}
\subcaption{Classification accuracy}
\end{minipage}%
\caption{Accuracy and FID on \celeba{} using simulator-generated data
with different $\nnvariationdegree$ schedules ($\epsilon=10$). The ``base'' parameter schedule across iterations is $[1000,500,200,100,50,20]$. ``Small'' (``large'') means fixing $\nnvariationdegree$ to be the smallest (largest) value across iterations. For example, the unit under ``large'' indicates the results when we set $\nnvariationdegree=[1000,1000,1000,1000,1000,1000]$.  } \label{fig:ablation_gamma}
\end{figure*}

Following the \pe{} paper \cite{lin2023differentially}, we gradually decrease them across  iterations. The idea is that larger variations at the beginning help \simpe{}/\pe{} explore and find good seeds, while smaller variations are needed later for convergence. In our experiments, most are set to follow an arithmetic or geometric sequence across iterations (\cref{app:exp_details}) based on heuristics. In the following, we modify their values: for each hyperparameter, we either fix it to the largest or smallest value in its base sequence. \cref{fig:ablation_alpha_beta,fig:ablation_gamma} show that:
\begin{packeditemize}
\item Setting these hyperparameters to a fixed \emph{large} value is not ideal. For instance, setting a large text variation throughout greatly hurts FID and accuracy. Conversely, using a fixed \emph{small} value has a smaller impact.
\item The default parameters do not always yield the best result (e.g., setting $\nnvariationdegree$ to a fixed small value improves FID). This suggests that our main results in the paper could be improved with better hyperparameters.
\end{packeditemize}

\FloatBarrier

\subsection{When to Switch from the Simulator to the Foundation Model}

In \cref{sec:result_simpe_both}, the foundation model was introduced after iteration 1. We vary it to be either earlier (after iteration 0, using only simulators for the initial samples) or later (after iteration 2). The result is in \cref{tab:simulator_diffusion_switch_iter}. We can see that:
\begin{packeditemize}
    \item Our reported main results can be improved with better parameters. For instance, starting the foundation model after iteration 0 improves accuracy to 74.3\% (+1.6\%) and reduces FID to 14.2 (-0.8) at $\varepsilon=1$.
    \item Nevertheless, the results across different schedules are relatively close. In most cases, they outperform using either the foundation model or the simulator alone (see \cref{tab:celeba_avatar}), reinforcing the conclusions in \cref{sec:result_simpe_both}.
\end{packeditemize}

\begin{table}[t]
\centering
\begin{tabular}{l|cc|cc}
\toprule
{Diffusion Start} & {FID ($\varepsilon=1$)} & {Acc ($\varepsilon=1$)} & {FID ($\varepsilon=10$)} & {Acc ($\varepsilon=10$)} \\
\midrule
After iter 0 & 14.2 & 74.3 & 13.2 & 73.6 \\
After iter 1 & 15.0 & 72.7 & 11.9 & 78.1 \\
After iter 2 & 16.0 & 73.6 & 14.2 & 76.5 \\
\bottomrule
\end{tabular}
\caption{Comparison of FID and accuracy with different diffusion start iterations for $\varepsilon=1$ and $\varepsilon=10$.}
\label{tab:simulator_diffusion_switch_iter}
\end{table}

\FloatBarrier

\subsection{Range of Simulator Parameters}

We vary the range of all numerical parameters in the \mnist{} experiment $\varepsilon=1$, including font size, rotation, and stroke width. Results across all \simpe{} iterations are presented.

\begin{packeditemize}
    \item \textbf{Font size.} In our main experiments, we used the range [10, 30). We evaluate two alternatives: [15, 25) (narrower) and [5, 35) (wider). See \cref{tab:font_size_range} for the results.
    \item \textbf{Rotation degree.}
In our main experiments, we used the range [–30, 30]. We evaluate two alternatives: [–15, 15] (narrower) and [–60, 60] (wider). See \cref{tab:rotation_degree_range} for the results.
    \item \textbf{Stroke width.} In our main experiments, we used the range [0, 2]. We evaluate two alternatives: [0, 1] (narrower) and [0, 3] (wider). See \cref{tab:stroke_width_range} for the results.
\end{packeditemize}

We can see from the above three experiments that:

\begin{packeditemize}
    \item For font size and rotation, varying the parameter range has minimal effect on accuracy or FID across iterations. In contrast, a suboptimal stroke width range ([0, 3]) leads to poor initial samples (FID 109.1), but \simpe{} gradually improves quality, narrowing the final FID gap to the best setting to 4.3.
    \item Regardless of the specific parameter range chosen, the overall trend is consistent: \simpe{} consistently improves the synthetic data over iterations and outperforms standard \pe{} significantly (see Table 1). These are the core messages of the paper, and they hold even when the parameter ranges are varied.
\end{packeditemize}

\begin{table}[th]
\centering
\begin{tabular}{l l |c c c c c}
\toprule
Range & Metric & Iter=0 & Iter=1 & Iter=2 & Iter=3 & Iter=4 \\
\hline
[15, 25) & Accuracy & 13.6 & 71.6 & 81.4 & 87.4 & 84.8 \\
         & FID      & 88.1 & 61.8 & 45.3 & 31.1 & 22.4 \\
\hline
[10, 30) & Accuracy & 16.6 & 76.7 & 86.7 & 89.5 & 89.1 \\
         & FID      & 85.8 & 57.0 & 42.4 & 28.7 & 20.7 \\
\hline
[5, 35)  & Accuracy & 13.3 & 66.0 & 82.8 & 90.6 & 86.2 \\
         & FID      & 87.2 & 58.2 & 41.4 & 28.7 & 20.2 \\
\bottomrule
\end{tabular}
\caption{Effect of font size range on accuracy and FID.}
\label{tab:font_size_range}
\end{table}

\begin{table}[th]
\centering
\begin{tabular}{l l |c c c c c}
\toprule
Range & Metric & Iter=0 & Iter=1 & Iter=2 & Iter=3 & Iter=4 \\
\hline
[–15, 15] & Accuracy & 14.5 & 75.7 & 86.0 & 88.3 & 87.7 \\
           & FID      & 83.1 & 56.7 & 42.3 & 28.6 & 19.7 \\
\hline
[–30, 30] & Accuracy & 16.6 & 76.7 & 86.7 & 89.5 & 89.1 \\
           & FID      & 85.8 & 57.0 & 42.4 & 28.7 & 20.7 \\
\hline
[–60, 60] & Accuracy & 13.3 & 72.3 & 83.0 & 88.5 & 86.3 \\
           & FID      & 86.6 & 56.0 & 40.9 & 27.5 & 19.4 \\
\bottomrule
\end{tabular}
\caption{Effect of rotation degree range on accuracy and FID.}
\label{tab:rotation_degree_range}
\end{table}

\begin{table}[th]
\centering
\begin{tabular}{l l |c c c c c}
\toprule
Range & Metric & Iter=0 & Iter=1 & Iter=2 & Iter=3 & Iter=4 \\
\hline
[0, 1] & Accuracy & 12.8 & 70.0 & 87.6 & 91.5 & 91.3 \\
       & FID      & 69.0 & 48.7 & 35.3 & 25.1 & 18.1 \\
\hline
[0, 2] & Accuracy & 16.6 & 76.7 & 86.7 & 89.5 & 89.1 \\
       & FID      & 85.8 & 57.0 & 42.4 & 28.7 & 20.7 \\
\hline
[0, 3] & Accuracy & 20.4 & 71.6 & 75.4 & 84.6 & 85.2 \\
       & FID      & 109.1 & 67.6 & 47.4 & 31.3 & 22.4 \\
\bottomrule
\end{tabular}
\caption{Effect of stroke width range on accuracy and FID.}
\label{tab:stroke_width_range}
\end{table}

\FloatBarrier
\subsection{Number of Samples Generated from the Simulator.}

We fixed the number of images to 60,000 across all experiments and iterations, following the DPImageBench evaluation protocol \cite{dpimagebench} for fair and consistent comparison across methods. Additionally, we run \mnist{} experiments ($\varepsilon=1$) with 30,000 and 120,000 samples in \cref{tab:num_samples_fid_acc}.

Consistent with the claims in the original \pe{} paper (Appendix N in \citet{lin2023differentially}), we see that more samples don’t always improve results. On one hand, more samples increase the chance of generating examples close to the private data, potentially improving synthetic data. On the other hand, a larger sample size flattens the DP Nearest Neighbors Histogram, reducing its signal-to-noise ratio and possibly degrading quality. The overall effect is nuanced. Following \citet{lin2023differentially}, we recommend setting the number of generated samples to be close to the number of private samples.

\begin{table}[th]
\centering
\begin{tabular}{l | c c}
\toprule
{\#Samples} & {FID} & {Accuracy} \\
\midrule
30,000  & 14.4 & 87.2 \\
60,000  & 20.7 & 89.1 \\
120,000 & 32.9 & 85.4 \\
\bottomrule
\end{tabular}
\caption{Performance comparison with different numbers of samples.}
\label{tab:num_samples_fid_acc}
\end{table}

\FloatBarrier

\subsection{The Impact of Distribution Alignment}

On \celeba{}, we show that \simpe{} with a simulator that aligns well with the private data outperforms standard \pe{} (\cref{sec:result_simpe_data,tab:acc_and_fid}). However, \simpe{} with a simulator less aligned with the private data performs worse than \pe{} (\cref{sec:result_simpe_both,tab:celeba_avatar}).

To provide a more controlled experiment, we fix the simulator type and only vary the alignment between the simulator and private data.

\begin{figure}[h]
    \centering
    \includegraphics[width=0.3\linewidth]{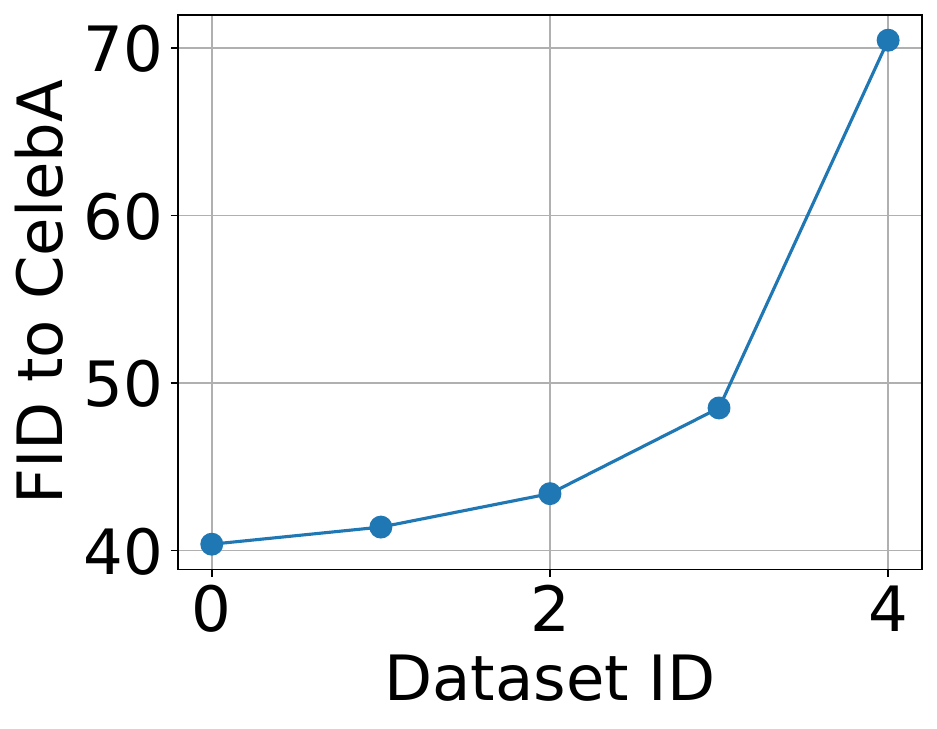}
    \caption{FID between the 5 constructed subsets of \digiface{} and \celeba{}.}
    \label{fig:abalation_datape_distribution_distance_data_construction}
\end{figure}

\myparatightestn{Experiment settings.} We use the same 1.2M simulator-generated images from a computer graphics renderer as in \cref{sec:result_simpe_data} and artificially adjust their alignment to \celeba{}. Specifically, for each image, we compute its distance to the closest \celeba{} image in the Inception embedding space, sort all images by this distance, and divide them into five subsets, $D_0,...,D_4$ (e.g., the closest 0.24M images form $D_0$). \cref{fig:abalation_datape_distribution_distance_data_construction} confirms that the FID between $D_i$ and \celeba{} increases with $i$, meaning that $D_0$ is the most aligned and $D_4$ the least. We then apply \simpe{} to each $D_i$ independently. 

\begin{figure*}[h]
\centering
\begin{minipage}{0.49\textwidth}
  \centering
\includegraphics[width=0.6\textwidth]{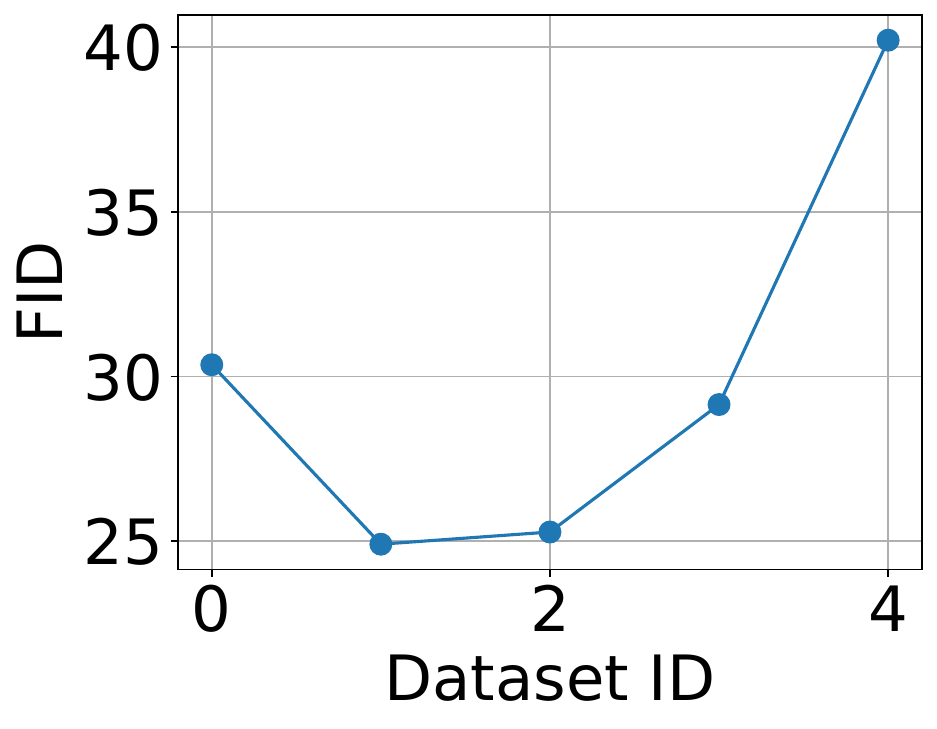}
\subcaption{FID}
\end{minipage}%
\begin{minipage}{0.49\textwidth}
  \centering
\includegraphics[width=0.6\textwidth]{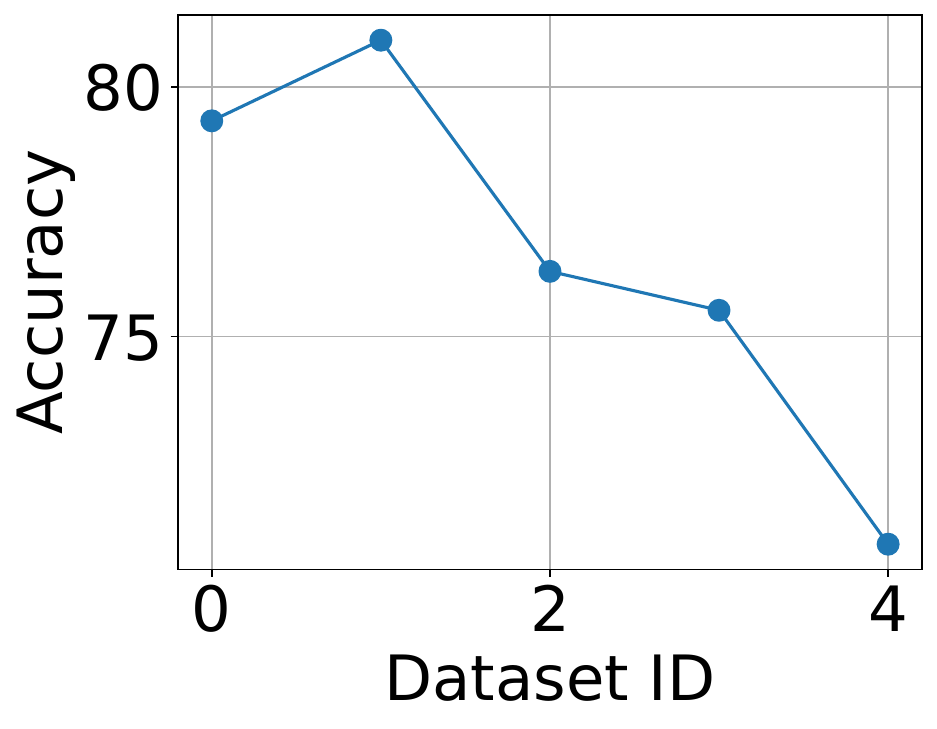}
\subcaption{Classification accuracy}
\end{minipage}%
\caption{Results of \simpe{}-generated data using the subsets in \cref{fig:abalation_datape_distribution_distance_data_construction} ($\epsilon=10$).  } \label{fig:ablation_datape_distribution_distance}
\end{figure*}

\myparatightestn{Results.} \cref{fig:ablation_datape_distribution_distance} shows that as alignment decreases, the sample quality generally drops. Specifically, \simpe{} with $D_0, ..., D_3$ yields better classification accuracy than \pe{}, while $D_4$ yields worse results than \pe{}. This confirms that \simpe{}’s performance is influenced by the degree of alignment. Additionally, in all cases, \simpe{}’s FID is better than the original simulator data’s FID, demonstrating \simpe{}’s ability to select useful samples.

\subsection{Class Label Information from the Simulators}
\label{app:class_label_info_from_simulators}

For simulator 1, the target class label (i.e., the digit) is fully controlled by one parameter.
For simulators 2 and 3, the target class label (i.e., the gender) is not directly controlled by any parameter, but could potentially be obtained by an external image gender classifier.
One benefit of using domain-specific simulators is that we can potentially use the class label information to enhance data quality \cite{wood2021fake,bae2023digiface}.
To get a more comprehensive understanding of \simpe{}, we consider two settings: \textbf{(1) Class label information is unavailable (abbreviated as ``\classunavail{}'').} We artificially make the problem more challenging by assuming that the class label information is \emph{not} available. Therefore, \simpe{} has to learn to synthesize images with the correct class by itself. 
Our main experiments are based on this setting.
\textbf{(2) Class label information is available  (abbreviated as ``\classavail{}'').} On \mnist{}, we further test how \simpe{} can be improved if the class label information is available. In this case, 
the \randomsampleapiname{} and \samplevariationapiname{} (\cref{eq:simulator_random_api,eq:simulator_variation_api}) are restricted to draw parameters from the corresponding class (i.e., the digit is set to the target class).

\begin{figure*}[!t]
\centering
\begin{minipage}{0.32\textwidth}
  \centering
\includegraphics[width=0.8\textwidth]{fig/5_mnist_images/image_sample/000000000.png}
\subcaption{Real (private) images}
\end{minipage}%
\begin{minipage}{0.32\textwidth}
  \centering
\includegraphics[width=0.8\textwidth]{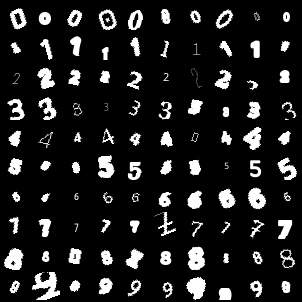}
\subcaption{Simulator-generated images}\label{fig:mnist_know_digit_simulator}
\end{minipage}%
\begin{minipage}{0.32\textwidth}
  \centering
\includegraphics[width=0.8\textwidth]{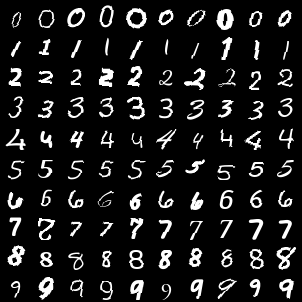}
\subcaption{\simpe{} images ($\epsilon=10$)}\label{fig:mnist_know_digit_simpe}
\end{minipage}
\vspace{-0.2cm}
\caption{The real and generated images on \mnist{} under the ``\classavail{}'' setting. Each row corresponds to one class.
The simulator generates images that are very different from the real ones. Starting from these bad images, \simpe{} can effectively guide the generation of the simulator towards high-quality images that are more similar to real data.} \label{fig:mnist_know_digit}
\end{figure*}

\myparatightestn{Results: Class label information from the simulators can be helpful.} %
The results are presented in \cref{tab:mnist_know_digit,fig:mnist_know_digit}. We observe that with digit information, the simulator-generated data achieves much higher classification accuracy (92.2\%), although the FID remains low due to the generated digits exhibiting incorrect characteristics (\cref{fig:mnist_know_digit_simulator}). The fact that \simpe{} outperforms the simulator in both FID and classification accuracy across all settings suggests that \simpe{} effectively incorporates private data information to enhance both data fidelity and utility, even when compared to such a strong baseline. As expected, \simpe{} under \classavail{} matches or surpasses the results obtained in \classunavail{} across all settings, suggesting the usefulness of  leveraging class label information.

\begin{table}[!t]
\small
    \centering
    \caption{
    Accuracy (\%) of classifiers trained on synthetic images and FID of synthetic images on \mnist{} under the ``\classavail{}'' setting. See \cref{tab:acc_and_fid} for results under the ``\classunavail{}'' setting for reference.}
    \label{tab:mnist_know_digit}
    \vspace{-0.1cm}
    \resizebox{0.5\linewidth}{!}{
    \begin{tabular}{l|cc|cc}
    \toprule
    \multirow{2}{*}{Algorithm} & \multicolumn{2}{c|}{{FID $\downarrow$}} & \multicolumn{2}{c}{{Classification Acc. $\uparrow$}}\\
    \cline{2-5}
     & $\epsilon = 1$ & $\epsilon = 10$ &  $\epsilon = 1$ & $\epsilon = 10$  \\
     \hline
    Simulator & \multicolumn{2}{c|}{86.0 ($\epsilon=0$)} & \multicolumn{2}{c}{92.2 ($\epsilon=0$)} \\
    \simpe{} & \textbf{20.7} & \textbf{8.6} & \textbf{93.9} & \textbf{95.5} \\
    \bottomrule
\end{tabular}}
\vspace{-0.3cm}
\end{table}
\FloatBarrier
\section{Efficiency Evaluation}
\label{app:efficiency}

As simulators could be much cheaper to generate samples than the foundation models, we show in this section that \emph{\simpe{} is much more efficient than \pe{}} in our experiments.

Note that the only difference between \simpe{} and \pe{} is the \randomsampleapiname{} and \samplevariationapiname{}. Therefore, we focus on comparing the computation time and peak CPU/GPU memory of the APIs. 
\cref{tab:efficiency_mnist,tab:efficiency_celeba_datape,tab:efficiency_celeba_avatar} show that \simpe{} APIs require far less time than \pe{}. For instance, on \mnist{}, each \pe{}’s API call takes over \emph{2400 GPU seconds}, whereas \simpe{} takes less than \emph{30 CPU seconds}--an 80x speedup, not to mention the lower cost of CPU than GPU. Consequently, each \simpe{} iteration is much faster than \pe{} (\cref{tab:iteration_run_time}). The only \simpe{} operation requiring GPU is computing the embedding and nearest neighbors of simulator-generated data (\cref{sec:method_data}), which is a one-time cost per dataset. Even this one-time process is significantly faster than one \pe{} API call. \simpe{} with data access requires more CPU memory to store simulator-generated data, but this can be easily optimized by loading only the needed data.

\begin{table}[h]
    \centering
    \resizebox{0.8\linewidth}{!}{
    \begin{tabular}{|c|c|c|c|c|}
        \hline
        & & \textbf{Time} & \textbf{Peak CPU Memory} & \textbf{Peak GPU Memory} \\
        \hline
        \multirow{2}{*}{\textbf{\pe{}}} & RANDOM\_API & 3920.30 seconds (GPU) & 15292.98 MB & 13859.10 MB \\
        & VARIATION\_API & 2422.44 seconds (GPU) & 15880.25 MB & 16203.89 MB \\
        \hline
        \multirow{2}{*}{\textbf{Sim-\pe{}}} & RANDOM\_API & 27.17 seconds (CPU) & 758.90 MB & 0 MB \\
        & VARIATION\_API & 18.51 seconds (CPU) & 1083.11 MB & 0 MB \\
        \hline
    \end{tabular}}
    \caption{Efficiency comparison of \pe{} and \simpe{} on the \mnist{} dataset. Tested on a Linux server with \texttt{AMD EPYC 7V12 64-Core Processor} and one \texttt{NVIDIA RTX A6000 GPU}.}
    \label{tab:efficiency_mnist}
\end{table}

\begin{table}[h]
    \centering
    \resizebox{0.8\linewidth}{!}{
    \begin{tabular}{|c|c|c|c|c|}
        \hline
        & & \textbf{Time} & \textbf{Peak CPU Memory} & \textbf{Peak GPU Memory} \\
        \hline
        \multirow{2}{*}{\textbf{\pe{}}} & RANDOM\_API & 39272.33 seconds (GPU) & 15293.29 MB & 13859.10 MB \\
        & VARIATION\_API & 31028.24 seconds (GPU) & 15879.47 MB & 16203.89 MB \\
        \hline
        \multirow{2}{*}{\textbf{Sim-\pe{}}} & RANDOM\_API & 0.04 seconds (CPU) & 61416.62 MB & 0 MB \\
        & VARIATION\_API & 0.03 seconds (CPU) & 61427.29 MB & 0 MB \\
        \hline
        \multicolumn{2}{|c|}{\textbf{Sim-\pe{} Setup}} & 1450.94 seconds (GPU) & 61420.69 MB & 3728.16 MB \\
        \hline
    \end{tabular}}
    \caption{Efficiency comparison of \pe{} and \simpe{} (with data access) on the \celeba{} dataset. Tested on a Linux server with \texttt{AMD EPYC 7V12 64-Core Processor} and one \texttt{NVIDIA RTX A6000 GPU}.}
    \label{tab:efficiency_celeba_datape}
\end{table}

\begin{table}[h]
    \centering
    \resizebox{0.8\linewidth}{!}{
    \begin{tabular}{|c|c|c|c|c|}
        \hline
        & & \textbf{Time} & \textbf{Peak Memory} & \textbf{Peak GPU Memory} \\
        \hline
        \multirow{2}{*}{\textbf{\pe{}}} & RANDOM\_API & 39272.33 seconds (GPU) & 15293.29 MB & 13859.10 MB \\
        & VARIATION\_API & 31028.24 seconds (GPU) & 15879.47 MB & 16203.89 MB \\
        \hline
        \multirow{2}{*}{\textbf{Sim-\pe{}}} & RANDOM\_API & 46.69 seconds (CPU) & 811.72 MB & 0 MB \\
        & VARIATION\_API & 48.78 seconds (CPU) & 1067.93 MB & 0 MB \\
        \hline
    \end{tabular}}
    \caption{Efficiency comparison of \pe{} and \simpe{} (with code access) on the \celeba{} dataset. Tested on a Linux server with \texttt{AMD EPYC 7V12 64-Core Processor} and one \texttt{NVIDIA RTX A6000 GPU}.}
    \label{tab:efficiency_celeba_avatar}
\end{table}

\begin{table}[h]
    \centering
    \resizebox{0.8\linewidth}{!}{
    \begin{tabular}{|c|c|c|c|}
        \hline
& \textbf{\mnist{}} & \textbf{\celeba{}} (\simpe{} with data access) & \textbf{\celeba{}} (\simpe{} with code access) \\
        \hline
        \textbf{\pe{}} & 22243.32 & 279695.52 & 279695.52 \\
        \textbf{Sim-\pe{}} & 607.95 & 441.63 & 880.38 \\
        \hline
    \end{tabular}}
    \caption{The runtime (seconds) of one iteration in \pe{} and \simpe{}. Tested on a Linux server with \texttt{AMD EPYC 7V12 64-Core Processor} and one \texttt{NVIDIA RTX A6000 GPU}.}
    \label{tab:iteration_run_time}
\end{table}
 
\FloatBarrier
\section{Extended Discussions}

\subsection{The Prevalence and Importance of Simulators}
\label{app:simulators}

In this paper, we define \emph{simulators} as any data synthesizers that do not rely on neural networks. These simulators are widely used across various applications due to their unique advantages over neural network-based approaches.

\myparatightestn{The prevalence of simulators.}
Despite the widespread adoption of foundation models, simulators remain highly prevalent. Here are a few notable examples:
\begin{packeditemize}
\item \textbf{Genesis:}\footnote{\url{https://genesis-embodied-ai.github.io/}} A physics-based simulator used in robotics, embodied AI, and physical AI applications. Since its release in late 2024, it has received over 24k GitHub stars and 84k downloads as of April 2024.
\item \textbf{Blender:}\footnote{\url{https://github.com/blender/blender}} A rendering framework widely used for image and video production, including in movie-making (see examples at Blender Studio\footnote{\url{https://studio.blender.org/films/}}). One simulator used in our face experiment (the \digiface{} dataset) is built on Blender.
\item \textbf{Unreal:}\footnote{\url{https://www.unrealengine.com/en-US}} A widely adopted game engine with image/video rendering capability. It holds a 14.85\% market share in the game development industry.\footnote{\url{https://6sense.com/tech/game-development/unreal-engine-market-share}} Note that competing engines also qualify as ``simulators'' under our definition.
\end{packeditemize}

\myparatightestn{The importance of simulators.}
As demonstrated above, simulators continue to play a crucial role in industry applications. Even synthetic data generation also frequently relies on simulators rather than foundation models (e.g., \metasim{} \cite{kar2019meta}, \digiface{} \cite{bae2023digiface}). This preference stems from several unique advantages of simulators over foundation models:
\begin{packeditemize}
\item \textbf{Rich annotations}: Simulators provide additional structured labels due to explicit control over the data generation process. For instance, besides face images, \facesynthetics{} \cite{wood2021fake} offers pixel-level segmentation masks (e.g., identifying eyes, noses), which are highly valuable for downstream tasks such as face parsing.

\item \textbf{Task-specific strengths}: Certain generation tasks remain challenging for foundation models. For example, despite recent advances, foundation models still struggle with generating images containing text \cite{rombach2022high}, whereas simulators can handle it easily. Our \mnist{} experiment was specifically designed to highlight this distinction.
\item \textbf{Domain-specific strengths}: In domains like networking, where robust foundation models are lacking, network simulators such as ns-3\footnote{\url{https://www.nsnam.org/}} remain a more reliable and scalable solution.
\item \textbf{Greater reliability and control}: Because simulators model the data generation process, they mitigate issues such as generating anatomically incorrect images (e.g., faces without noses\footnote{\url{https://www.reddit.com/r/StableDiffusion/comments/1e7dd62/weird_and_distorted_images_with_a1111_sd3/}}).

\end{packeditemize}

That said, there are indeed data domains for which no suitable simulators exist. However, the goal and contribution of this work is not to cover every possible data domain--which is inherently infeasible--but rather to significantly expand the range of domains where \pe{} can be applied. For example, we demonstrate its effectiveness on challenging domains for \pe{} such as text rendering and face generation.

Moreover, the proposed approach is not limited to using simulators. The method in \cref{sec:method_data} can also be applied to any public dataset, as demonstrated in \cref{app:cifar10}. This highlights the broader potential of our approach that sidesteps the limitation of requiring a suitable simulator.

\subsection{Discussion on the Chosen Metrics}
\label{app:metrics}

In the main experiments, we use \textbf{FID} and \textbf{the accuracy of downstream classifiers} as the metrics.

\myparatightestn{Details on the accuracy of downstream classifiers.} 
To generate synthetic data, we use the conditional version of \pe{} (\cref{app:pe}), 
so that each generated images are associated with the class labels (i.e., `0'-`9' digits in \mnist{}, male vs. female in \celeba{}). 
These class labels are the targets for training the classifiers.
We employ a strict train-validation-test split and account for the privacy cost of classifier hyperparameter selection. Specifically, we divide the private dataset into disjoint training and validation sets. We then run \simpe{} on the training set to generate synthetic data. Next, we train three classifiers—ResNet \cite{he2016deep}, WideResNet \cite{zagoruyko2016wide}, and ResNeXt \cite{xie2017aggregated}—on the synthetic data and evaluate their accuracy on the validation set. Since the validation set is part of the private data, we use the Report Noisy Max algorithm \cite{dwork2014algorithmic} to select the best classifier checkpoint across all epochs of all three classifiers. Finally, we report the accuracy of this classifier on the test set.
This procedure ensures that the reported accuracy is not inflated due to train-test overlap or DP violations in classifier hyperparameter tuning. 

\myparatightestn{Why we pick these metrics.} We chose FID and classification accuracy because they are the most widely used metrics in DP image synthesis. \emph{Most of the baselines we compared to utilize only these two metrics or just one of them \cite{dpimagebench}}. The only exception is GS-WGAN, which also uses Inception Score (IS) \cite{salimans2016improved}. However, IS is more relevant for natural images (e.g., ImageNet images) because it evaluates the diversity across ImageNet classes, making it unsuitable for the datasets we considered in the main paper. Additionally, for the downstream task (ten-class classification for MNIST and binary classification for CelebA), we also follow prior works \cite{dpimagebench}. 

\myparatightestn{The differences between these metrics.} FID measures the ``fidelity'' of synthetic data by mapping both synthetic and private data to an embedding space, approximating each with a Gaussian distribution, and calculating the Wasserstein distance between the two distributions. Classification accuracy measures the ``utility'' of synthetic data by evaluating its performance when used to train a downstream classifier, simulating real-world use where users expect the synthetic data to support good classifier performance on real data. \emph{FID and classification accuracy are complementary}: FID focuses on distribution-level closeness, while classification accuracy is sensitive to outliers or samples near the classification boundary. A high score in one does not necessarily correlate with a high score in the other. This is why we often see cases where the best methods for these metrics differ \cite{dpimagebench}.

\subsection{Bias in Simulators}
Because \simpe{} relies on simulators, biases inherent in the simulator can be transferred to the synthetic data produced by \simpe{}.

\subsection{The Ordering of Applying the Simulator and the Foundation Model}
\label{app:simulator_and_model_order}

In \cref{sec:method_simulator_and_model}, we propose to use simulators in early iterations and then switch to foundation models. The reverse ordering--``foundation model $\to$ simulator''--is also feasible with appropriate modifications. In this setup, when transitioning from the foundation model to the simulator, one would need to find the optimal simulator parameters corresponding to each sample generated by the foundation model, potentially using gradient-free optimization methods.

However, we only explore ``simulator $\to$ foundation model'' in the paper for two main reasons:

\begin{packeditemize}
    \item This ordering keeps the approach simple and avoids the need for special treatments like the one described above.
    \item It is also more promising in data quality. Simulators can generate diverse samples that span a wide range of variations, making them ideal for initialization. However, their image quality is bad. Foundation models, on the other hand, are better at producing high-fidelity, realistic samples, which is more desirable in the final stages of the pipeline, which creates the final synthetic data.
\end{packeditemize}

\end{document}